%% file: iclr2025_conference.tex
\pdfoutput=1
\documentclass{article} %
\usepackage{iclr2025_conference,times}

\input{math_commands.tex}

\usepackage[hidelinks]{hyperref}
\usepackage{url}
\usepackage{booktabs}
\usepackage{algorithm}
\usepackage{algpseudocode}
\usepackage{tikz}
\usepackage[scaled=0.92]{helvet}   %
\usepackage{pgfplots}
\usepackage{listings}
\pgfplotsset{compat=1.16}
\usetikzlibrary{arrows.meta,positioning,fit,backgrounds,patterns}
\usepgfplotslibrary{groupplots}

\usepackage{enumitem}
\setlist{topsep=2pt,itemsep=2pt,parsep=0pt,leftmargin=1.4em}
\title{A Self-Improving Agent that Grows its Own\\ DSL: Oracle-Free Operator Induction\\ for Exact Question Answering}

\author{%
Yun-Shiuan Chuang\textsuperscript{*}, Yirou Ge\textsuperscript{*}, Yixi Li\textsuperscript{*}, Yiming Bian, Zijie Pan,\\[3pt]
\textbf{Jenna L.\ Ballard, Uma Kona, Linsey Pang, Prakhar Mehrotra}\\[6pt]
PayPal AI}

\newcommand{\opset}{\ensuremath{\mathcal{O}}}          %
\newcommand{\clos}[1]{\ensuremath{\mathrm{cl}(#1)}}    %
\newcommand{\recs}{\ensuremath{M}}                     %
\newcommand{\vsig}[1]{\ensuremath{\sigma(#1)}}         %

\iclrfinalcopy
\begin{document}

\maketitle
{\renewcommand{\thefootnote}{\fnsymbol{footnote}}\footnotetext[1]{Equal contribution.}}

\input{sections/abstract}

\input{sections/intro}
\input{sections/preliminaries}
\input{sections/methods}
\input{sections/exp_setup}
\input{sections/results}
\input{sections/related_work}
\input{sections/conclusion}

\input{sections/repro_statement}

\input{sections/ethics_statement}

\input{sections/llm_usage_statement}

\input{iclr2025_conference.bbl}
\appendix
\input{sections/dataset}
\input{sections/appendix}

\end{document}

%% file: math_commands.tex
\usepackage{amsmath,amsfonts,bm}

\def\eqref#1{equation~\ref{#1}}
\def\1{\bm{1}}

\DeclareMathAlphabet{\mathsfit}{\encodingdefault}{\sfdefault}{m}{sl}
\SetMathAlphabet{\mathsfit}{bold}{\encodingdefault}{\sfdefault}{bx}{n}

%% file: sections/abstract.tex
\begin{abstract}
A self-improving agent has to decide which new skills to keep, and in practice it keeps whatever earns a high
reward, convinces a judge model, or compresses its library, even when the skill is wrong. That is fine until the
answers have to be exactly right: over a table of transactions, a total is either right or wrong. We ask a
harder question: can an agent grow what it can compute with no reward, no judge, and no answer key, and still
be exactly right? A single frozen model runs two nested loops over a small, exact \emph{domain-specific
language} (DSL). The inner loop answers new questions only when two independently written expressions agree on
every record. The outer loop discovers its own missing capability and authors an operator for it, admitted only when an
independent second implementation agrees and the result obeys a conservation law, all without ground truth. The inner loop roughly triples the
answerable-question set, and the outer loop self-extends on disjoint starting sets, inventing operators outside the
starting closure; a no-induction control answers none of the new questions. The admission rule is empirically
sound: it rejects 138 of 138 buggy operator variants under an independent reference, even one written by a
different model family, where the admit-if-it-runs, self-consistency, and compression signals the field relies on
admit most of them. Unmodified, the same mechanism self-extends on two further domains with unrelated schemas,
business-ledger and clinical-encounter question answering, growing their answerable sets more than tenfold and
nearly twentyfold. The loop transfers across model families too: run with further proposers, the gate stays sound
and independent families self-extend comparably, with only the induction yield tracking each proposer's DSL
fluency.
\end{abstract}

%% file: sections/intro.tex
\section{Introduction}

Getting a language-model agent to improve itself is easy to want and hard to trust. An agent can fine-tune its
weights, rewrite its prompts, or grow a library of skills~\citep{wang2024voyager, cai2024large}, and in each case it keeps a new capability because
that capability earned a high reward~\citep{romera2024mathematical, novikov2025alphaevolve}, convinced a judge model, or compressed the library~\citep{ellis2021dreamcoder, grand2024lilo, wang2024trove}. With no ground truth,
those are about the best signals available. When the answers have to be exactly right, they are not enough. A
wrong skill can score well, satisfy a judge, and compress the library as neatly as a correct one. Ask ``how much
did my top buyer spend?'' of a merchant's transaction history and the total is either right or wrong. We study
self-improvement with no teacher: can an agent expand what it can compute using no reward, no judge, and no
answer key, while every answer stays exactly checkable by a trusted program?

One dependable route to exact answers is to have the model write a program instead of doing the arithmetic in
its head. Program-of-thoughts methods \citep{chen2023pot} take this route. In our setting the program is an
expression in a small sandboxed \emph{domain-specific language} (DSL): a set of operators that compose like code
and run under a trusted interpreter. But coverage is the limit. A fixed operator set \opset{} answers exactly the
questions you can build by composing its operators, its closure \clos{\opset}. A per-group total, when the DSL has only whole-table reductions, is simply
out of reach, and writing a fresh operator by hand for every question a domain raises does not scale. So the
agent has to grow its own DSL, adding the operators its questions demand, without ever admitting one it cannot
verify. Think of a language with a fixed vocabulary: you can only say what its words
combine to express, so the agent has to coin new words. A coined word is dangerous if no one can check what it
means.

Whether this works then depends on the rule for admitting a new operator. A rule built on reward, compression,
or a model's judgment does not
preserve exactness: a wrong operator can score well, compress beautifully, or convince a judge, and even a judge
that happens to be right supplies no checkable guarantee (Appendix~\ref{app:judge}). We hold a harder
line: an operator is admitted only on evidence that
needs no answer key and no judge. The model writes it twice, independently, and the two implementations must
agree on every record. It must also obey conservation laws stated over the operators already trusted, the
simplest being that per-group totals add back up to the global total. This rule is oracle-free but
\emph{empirical}, not a formal proof: independent agreement is checked on a finite validation set within a small
numerical tolerance, and the conservation laws are the ones the model states, so admission certifies an operator
against targeted tests rather than over all possible inputs.

The method is two nested loops (Figure~\ref{fig:loops}, Appendix~\ref{app:algo}), both run by a single frozen model. The inner loop adds
no operators. It proposes questions it has not seen, answers each with one expression, and keeps the answer only
when a second expression, written from the question alone, agrees on every record and computes something new. The
outer loop adds capability. When the model keeps asking for something outside \clos{\opset}, it works out what is
missing, writes an operator for it, and puts it through the gate above. If the operator passes, it joins the
library and the inner loop runs again on the larger set.

We make six contributions. First, a teacher-free basis for self-improvement: an agent grows what it can compute
using only evidence that needs no answer key, two independently written programs that agree on every record and
conservation laws over operators it already trusts, replacing the reward, judge, and compression signals that
self-evolving agents and library-learning systems~\citep{grand2024lilo, wang2024trove} rely on. Second, a two-loop,
single-model self-extending DSL agent that instantiates it, with direct evidence the admission rule is sound: it
admits none of 138 buggy operator variants under an independent reference, including ones built to slip past the
agreement check, and stays sound with the reference in a \emph{separate} call and from a \emph{different model
family}, Gemma-3-27B~\citep{team2025gemma}, whereas admit-if-it-runs, self-consistency, and compression each admit
most. Third, a \emph{gap-seeking} use of the frozen model that surfaces and ranks the capabilities its own
questions demand but its operators cannot express, so the agent picks what to add. Fourth, a controlled end-to-end
study on financial-transaction QA over a fully synthetic corpus (no real customer data; Appendix~\ref{app:dataset}):
the inner loop roughly triples coverage across 13 starting sets, the outer loop extends three disjoint sets into
operators outside the original closure with a no-operator control isolating them, and the grown DSL agrees exactly
with an independent human-authored gold extractor on every compute type. Fifth, evidence it is not specific to
merchant transactions: with no code changes beyond a schema-adapting parser, the same loop self-extends on two
further domains, business-ledger and clinical-encounter QA (Appendices~\ref{app:booksql},~\ref{app:ehrsql}). Sixth,
evidence it is not tied to one model family: run with three further proposers the gate stays sound for all, and two
independent families, Mistral-Small-24B~\citep{mistralai2025small3} and Llama-3.3-70B~\citep{llama2024herd, metallama2024llama33},
self-extend comparably, with yield tracking each proposer's DSL fluency (\S\ref{sec:model-ladder}).

%% file: sections/preliminaries.tex
\section{Preliminaries: the base agent and the exact-verifiability invariant}
\label{sec:prelim}
The base agent answers a question about one record sequence by emitting a single expression in the restricted
DSL. Expressions are parsed to an abstract syntax tree and evaluated by a sandboxed interpreter
that allows only a whitelisted set of node types and calls to bare, registered functions; there is no
attribute access, no arbitrary call target, and no state. Every operator is a small, audited Python function
with exact numeric semantics. The invariant we preserve throughout is: an answer is trusted only if it is
computed by this interpreter, and a new operator is trusted only if its correctness is established by
relations that hold independently of any single model's judgment.

\paragraph{Notation.} We validate expressions on a fixed set of records \recs{}. For an expression $e$, write
\vsig{e} for the vector of $e$'s values across \recs{}, its \emph{value signature}; two expressions with equal
signatures compute the same quantity on \recs{}. An operator set \opset{} answers a question when some expression
over \opset{} matches the question's value signature, and the set of all such questions is the closure
\clos{\opset}. The inner loop enlarges the answerable set within a fixed \clos{\opset}; the outer loop enlarges
\opset{} itself, and hence \clos{\opset}.

\paragraph{Fixed starting point.} Every experiment begins from one fixed, minimal state: a small set of seed
question types and \emph{exactly} the operators needed to answer them, and no others. That fixed start is the
control that makes ``extension'' meaningful: an operator the outer loop induces is genuinely new only if it lies outside
the closure of this fixed initial set, so we never leak a target capability into the starting library.

\paragraph{Substrate and its boundary.} The loop grows the operator vocabulary but never alters a fixed
\emph{substrate}: the raw data representation and an attribute-free, first-order sandbox for operator bodies
(Appendix~\ref{app:sandbox}). This keeps exact verifiability total, every induced operator composes over audited
base operators in a language too small to escape to arbitrary computation, so the library stays auditable with no
supervision. The substrate has a precise, checkable \emph{reach}; a gap needing something outside it (e.g.\ banned
attribute access) cannot be closed, and extending the reach is a deliberate, logged act, not a relaxation of the
invariant.

%% file: sections/methods.tex
\section{Method}
\label{sec:method}

\input{figures/two_loops}

\subsection{Inner loop: expanding coverage within a fixed operator set}
Given a starting set of question types and a fixed operator set, the inner loop (\textsc{InnerLoop} in
Algorithm~\ref{alg:selfextend}, Appendix~\ref{app:algo}) runs four stages, all by the frozen base model.
\begin{enumerate}
\item \textbf{Propose.} Shown a few example (question, expression) pairs, the model invents different natural
questions one expression over the same data can answer; we draw many samples.
\item \textbf{Execute.} Each expression is evaluated on every validation record (held-out synthetic merchants,
Appendix~\ref{app:dataset}) under an interpreter exposing \emph{only} the current operators; those calling an
outside operator or not returning a single number are discarded.
\item \textbf{Verify.} The model then answers the \emph{question text alone}, separately, and the proposal is kept
only if this independently written expression matches the first on every record, strong evidence they compute the
same quantity with no ground-truth answer.
\item \textbf{Keep if new.} A verified quantity counts only if its value across the records differs from every
starting type and every quantity already accepted.
\end{enumerate}
The inner loop adds no capability: everything recombines the current operators, measuring how much larger the
answerable set is than the starting set.

\subsection{Outer loop: inducing a new operator, verifiably}
Restricted to a small operator set, the base model repeatedly reaches the boundary of what it can express and
\emph{names} the missing capability. The outer loop (\textsc{OuterLoop} in Algorithm~\ref{alg:selfextend}) turns
such a recurring request into a new operator in four steps: (1) discover the gap by tallying which missing
capability its questions most demand, and author an operator; (2) validate it with no ground-truth answer;
(3) re-run the inner loop with the enlarged set; (4) report. Admission is gated, not teacher-selected, so the agent
attempts the top-$K$ gaps in rank order ($K=3$) and keeps the first admissible operator, so an un-inducible top
entry does not stall the loop, at no cost to soundness (Appendix~\ref{app:variance}).

The most-requested capability is per-group aggregation, not expressible by the 9 whole-table reducers. The model
authors \texttt{per\_group\_sum}$(keys, values)$, returning the per-key sum as a plain list, written
\emph{method-free} (bare calls and comprehensions over the seed reducers, no attribute or method calls) as the
sandbox of Section~\ref{sec:prelim} requires; a list keeps the result composable, so
\texttt{max(per\_group\_sum(buyer\_ids, amounts))} answers ``which buyer's total spend is highest'' as one
expression, unlocking a whole per-group family. A per-group count is admitted the same way (group counts sum to the
record count), and its frequency comes with it; the mode \emph{value}, a non-numeric argmax with no conservation
law, stays outside (Section~\ref{sec:mode-limitation}).

\paragraph{Validation with no answer key.} A brand-new operator has nothing to check against, so we admit it the
way one trusts a new instrument with no reference standard: build it twice, independently, and require the copies
to agree, and require it to obey conservation laws already trusted. Concretely, on the validation records and
fuzzed inputs (empty, singleton, duplicate-key, missing, permuted):
\begin{enumerate}
\item \emph{Two implementations agree}, i.e.\ differential testing~\citep{mckeeman1998differential, yang2011csmith}:
the model builds the operator twice and the two must return equal value signatures across the records (relative
tolerance $10^{-6}$). The reference is generated separately, shown only the question text and told to use a
different coding style; Section~\ref{sec:soundness} measures how much this independence buys, including when both
copies share a mistake.
\item \emph{Conservation and structure hold}, i.e.\ metamorphic relations~\citep{chen1998metamorphic, segura2016survey}
written only from already-trusted operators (e.g.\ per-group totals sum to the global total,
\texttt{sum(op(keys, amounts)) == sum(amounts)}; one result per distinct key; empty in, empty out; order
invariance). Each relation must actually bind a trusted operator (a relation that is just \texttt{True} is
rejected) and is checked on populated validation data; edge cases such as missing keys are covered by the
differential fuzz inputs, so the gate declines rather than silently admits (Appendix~\ref{app:gate}).
\item \emph{The answers hold up downstream}: every question the operator unlocks must still pass the inner loop's
independent-agreement check.
\end{enumerate}
No single test is an answer key; a wrong operator would have to pass all of them together.

%% file: figures/two_loops.tex
\begin{figure}[t]
\centering
\includegraphics[width=\textwidth]{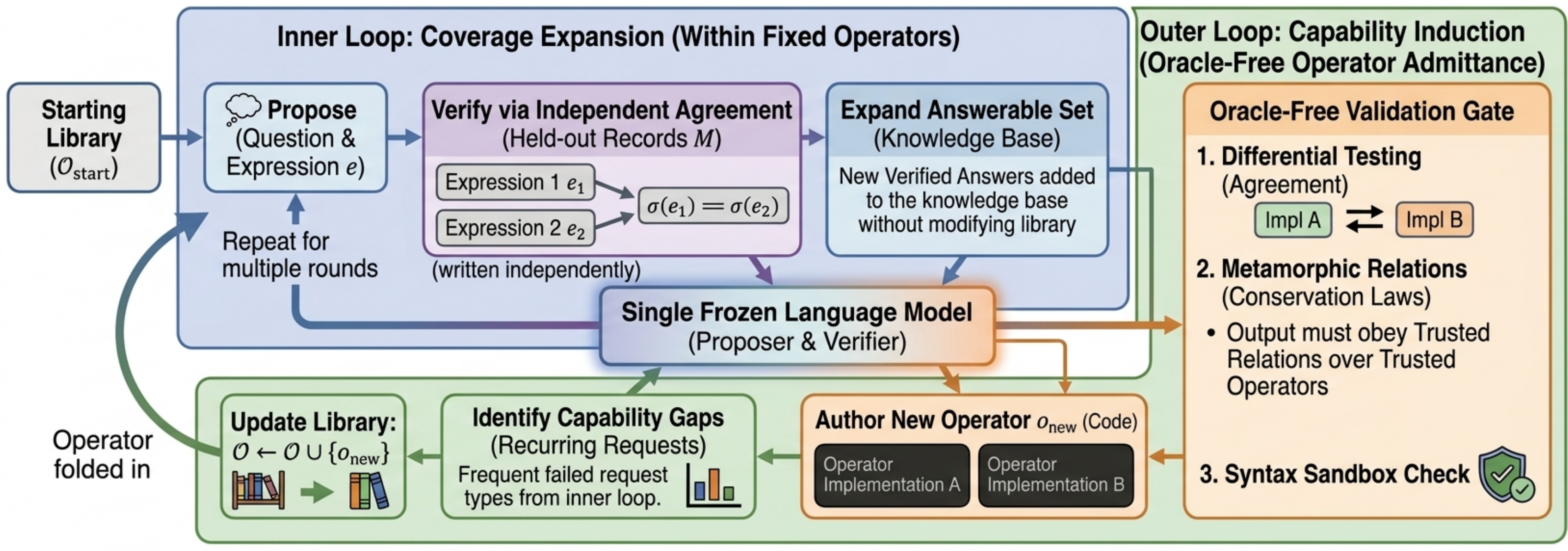}
\caption{The two nested loops, both run by a single frozen model with no ground-truth oracle and no judge. The
inner loop (blue) expands the answerable set within a fixed operator set, keeping a question only when two
independently written expressions agree on every record ($\sigma(e_1)=\sigma(e_2)$). The outer loop (green)
induces a new operator when a capability gap recurs, admits it through the oracle-free validation gate, and folds
it into the library before re-running the inner loop. Detailed in Section~\ref{sec:method}.}
\label{fig:loops}
\end{figure}

%% file: sections/exp_setup.tex
\section{Experiments}

\paragraph{Setup.} All runs use a single frozen mid-size open model, Qwen3.6-35B-A3B~\citep{qwen3.6-35b-a3b}, no
fine-tuning, reasoning off, temperature 1.0. Every merchant-transaction experiment uses a fully synthetic merchant
corpus (Appendix~\ref{app:dataset}), sampled from aggregate statistics of a real transaction distribution; no real
user or customer record is used anywhere. (The domain-transfer studies of \S\ref{sec:domain-transfer} instead use
BookSQL and EHRSQL over MIMIC-III; see Appendices~\ref{app:booksql},~\ref{app:ehrsql}.) Each candidate is validated
on a held-out set of 63 synthetic merchants (43 in the text-to-SQL study), held fixed across the trajectory run so
its curve is comparable end to end; the sampling budget is stated with each result and jobs run on B200 GPUs.
Appendix~\ref{app:repro} gives the environment and reproduction protocol, and Appendices~\ref{app:sandbox}
to~\ref{app:sql} detail the sandbox, prompts, seeds, and each experiment's protocol.

%% file: sections/results.tex
\paragraph{Overview.} Figure~\ref{fig:trajectory} shows the whole system running end to end on merchant
transactions and, with no code changes beyond a schema-adapting parser, on two further domains with unrelated
schemas: the inner loop expands coverage within a fixed operator set, an outer-loop induction adds operators
that the loop then uses to climb further, and the run stops on its own when the next gap has no checkable
invariant.

\input{figures/trajectory}

\subsection{Inner loop: coverage expansion}
\label{sec:inner}
From an amounts-and-timing starting set of 10 question types and 9 operators, one inner round proposes 82 candidate
questions, of which 40 run to a single number on every record, 35 pass the independent-expression check, and 26 are
new quantities, so the answerable set grows $10\to36$ with \emph{no new operator}. Repeating (counting only what
earlier rounds missed), the new counts are $26,18,9$ as the total climbs $36,54,63$ (Table~\ref{tab:saturation},
Appendix~\ref{app:seeds}); returns diminish, and a robust stopping rule is left to future work.

\paragraph{Robustness.} The expansion is not specific to one starting set: across 13 starting sets (3 hand-chosen,
10 random) every one yields $+9$ to $+39$ new verified types in a single round (mean $+18.3$), and across 8
sampling-budget settings the count never approaches zero.

\paragraph{Held-out validity against independent gold.} Coverage counts the loop's own round-trip-verified types,
so a fair worry is that it certifies internal consistency but not correctness. As an external check we score the
DSL's expressions against an \emph{independent} gold-answer extractor (the leaderboard gold pipeline, with
separately-implemented answer functions); both are deterministic code, so their agreement is a property of the
code, not the corpus. Across 3{,}000 held-out merchants the DSL and the gold agree exactly on every precise/compute
type on non-truncated sequences, the only systematic disagreement being the count-tie ``busiest day/hour''
categories (gold breaks ties by dollar volume, the DSL by first occurrence). The growth counts (e.g.\ $10\to336$)
thus measure the diversity of exactly-computable quantities this proposer reaches and verifies, and this check
confirms they are computed correctly.

\subsection{Outer loop: automated operator induction}
\label{sec:results}
The fully closed loop runs on three independent starting sets, each with the minimal operator set that answers its
seed types and each \emph{inducing from its own} operators (Appendix~\ref{app:seeds}): amounts/timing (9
operators), top-buyer (8), and timing (15). Each set \emph{discovers its own gap} by tallying which capabilities
its questions most demand outside its closure, then \emph{authors} operators to close it, admitted only on
evidence needing no answer key. The gap differs by set (amounts/timing and timing most-request a \emph{mode}
reducer, top-buyer a \emph{count}), but in every set it lies in the same family, per-group aggregation, that none
can express, and the loop \textbf{self-extends}: each set admits 2 to 6 distinct per-group operators with no
oracle (deduplicated by value signature). The workhorse is \texttt{per\_group\_sum}; a count gap instead admits a
per-group \emph{count} on the analogous conservation law. The gate rejects other drafts on the failure modes it
should (reverted method calls, disagreement, a false relation), and the admitted operators span several families,
not per-group alone (filtering, thresholding, binning, distinct counts; Appendix~\ref{app:families}).
Self-extension does not depend on co-generating the reference: with a separate-call reference each set still
self-extends, admission yield $14\pm7.4$, $2.4\pm0.8$, $4.6\pm1.5$ operators, positive in $5/5$ runs
(Appendices~\ref{app:independence},~\ref{app:variance}).

\paragraph{The induced operators are used, and a control isolates them.} Re-running the inner loop with the enlarged
set (Table~\ref{tab:closed}), a meaningful share of newly answerable types call an operator admitted this round:
$25$ of $65$ (amounts/timing), $6$ of $62$ (top-buyer), $2$ of $66$ (timing), all per-group totals, counts, or
derived statistics no composition of the starting operators could express. The control (same round, no new
operator) yields new types but \emph{zero} using any per-group operator, isolating induction as the cause.

\paragraph{The loop chains across rounds.} With round-1 operators in the library, the model discovers a
\emph{further} gap and induces more operators outside the enlarged closure ($2$, $4$, $1$ for amounts/timing,
top-buyer, timing, closing a weighted-mode, equality, and argmax gap), and uses them: cumulatively $116$, $254$,
and $53$ of each set's answerable types call an induced operator (Figure~\ref{fig:seed-trajectory},
Table~\ref{tab:closed}), so the extension \emph{compounds} rather than repeating.

\paragraph{Two further growth modes.} The three disjoint seeds converge on a shared answerable set, each computing
the same quantities through independently authored per-group operators admitted to identical values; and the same
discipline supports verified \emph{compression}, folding a recurring pattern into a named operator admitted only if
equivalent to the pattern on every record, unlike library-learning systems that refactor and trust the result
\citep{ellis2021dreamcoder} (Appendices~\ref{app:convergence},~\ref{app:compress}).

\subsection{Gate soundness: a mutation study}
\label{sec:soundness}
The gate admits operators with no answer key, so a fair worry is that it rubber-stamps a plausible but wrong one.
For each admitted operator we generate every single-point mutation of its implementation; a mutant that still
matches the original on every input is discarded as equivalent, leaving 138 genuinely buggy mutants across the 14
operators (the original labels mutants only; the gate never sees it). Each runs through the same gate under two
threat models (Appendix~\ref{app:mutation}, Table~\ref{tab:ablation}). With an independent reference, the setting
the loop uses, the false-admit rate is $0/138$: differential testing kills all 138. In the \emph{correlated}
worst-case, where the identical bug is written into both implementations so differential is blind, the metamorphic
conservation checks still leave only 10 admitted ($10/138$). The two layers are complementary, independent
agreement catches independent mistakes and conservation against trusted primitives catches shared ones, and the
residual $10/138$ arises only when independence is deliberately defeated, which motivates the protocol below.

\paragraph{Independence: the reference need not be co-generated.} The implementation and its reference are
written in one response, so a fair objection is that the differential is not a genuine independence test, a
shared error could sit in both. We re-run the mutation study with the reference generated in a \emph{separate}
call, conditioned only on the operator's name, signature, and one-line spec and never shown the implementation.
A fully independent reference still admits $0/138$ buggy mutants, matching the co-generated setting, so soundness
does not depend on co-generation. The reference need not even come from the same model: written by a different
family, Gemma-3-27B, it still admits $0/138$ (Appendix~\ref{app:independence}).

\paragraph{A gate component ablation.} Toggling each layer over the same pool (Table~\ref{tab:ablation}) shows the
two verification layers are complementary: differential testing kills all 138 buggy mutants under an independent
reference ($0/138$) but is defeated when the same bug is co-written into both implementations ($129/138$), the
regime the metamorphic conservation checks cover instead ($10/138$ on their own, and the full gate's $10/138$
correlated worst-case versus $0/138$ under the independent reference the loop uses); the method-free sandbox alone
admits $129/138$. Precision is high but not perfect: $1$ of the $14$ correct operators is rejected, by a mis-stated
relation rather than a wrong value, and the outcome is invariant to validation-set size ($0/138$ at $N=3$ to $63$;
Appendices~\ref{app:false-reject},~\ref{app:mutation}).

\input{tables/gate_ablation}

\paragraph{No teacher-free signal substitutes for the gate.} Two natural alternatives fail where the gate does not
(Table~\ref{tab:ablation}): admitting whatever runs (the no-oracle analog of library-learning admission) and
self-consistency each admit $129/138$ buggy operators and a reuse/compression rule $101/138$, while asking the
frozen model to judge correctness wrongly rejects 2 of 14 correct operators with none of the gate's checkable
basis (Appendices~\ref{app:admission-baselines},~\ref{app:judge}).

\paragraph{Adversarial constructions bound the guarantee.}\label{sec:adversarial} A bug \emph{designed with full
knowledge of what the gate checks} can pass: a conservation-preserving \emph{relabeling} of a per-group output
defeats every symmetric relation, but harmlessly, since the sandbox exposes no keyed lookup that could surface it;
and a size-changing merge escapes only when its witnessing relation names a bare builtin rather than a trusted
operator (the study's single false admit). Both are characterized rather than hidden, and re-validating with one
added call simulating keyed access catches every relabeling under an independent reference ($4/4$;
Appendix~\ref{app:adversarial}).

\subsection{The mechanism transfers across base models}
\label{sec:model-ladder}
Everything so far runs on one base model. To check self-extension is a property of the method, we re-run the entire
outer loop with four Qwen3 \emph{dense} models~\citep{yang2025qwen3} (4B/8B/14B/32B) across all three domains
(Figure~\ref{fig:model-ladder}); the 35B-A3B main model is a later-generation mixture-of-experts ($\approx$3B active
parameters), read as a cross-architecture point rather than the top of a size axis. The seed-sensitivity companion
across all three merchant seeds is in Figure~\ref{fig:seed-model-ladder} (Appendix~\ref{app:model-ablation}).

The gate stays sound at every scale on every domain: no wrong operator is admitted by any model, and where
induction fails the model simply wrote nothing admissible. Induction fires across the size range, not only at the
top, though not monotonically: every dense model except 8B self-extends on merchant amounts/timing, 32B on all
three merchant seeds, every dense model on BookSQL, and only 32B on EHRSQL. The 35B-A3B main model self-extends on
every merchant seed and both transfer domains despite far fewer active parameters than the largest dense model, so
what the gate gates on tracks capability, not active-parameter count or architecture (per-model counts in
Appendix~\ref{app:model-ablation}).

We read the ladder for gate soundness and whether induction fires, not as a coverage ranking: absolute height on
the metadata-rich BookSQL/EHRSQL schemas is volume-sensitive (a weaker model emitting more candidate expressions
accumulates more verified signatures), a bounded effect that does not reach the growth claims
(Appendix~\ref{app:model-ablation}).

\input{figures/base_model_ladder}

\paragraph{Three further model families.} To check the mechanism is not tied to Qwen, we re-run the entire closed
loop, same frozen prompt and gate, with three independent families as the proposer, Mistral-Small-24B, Gemma-3-27B,
and Llama-3.3-70B (Figure~\ref{fig:cross-family}, Table~\ref{tab:cross-family}). The gate stays sound for all four
families: no wrong operator is admitted by any family on any domain or seed. Induction transfers too: Mistral and
Llama both drive the loop to fresh coverage and induce sound operators (Mistral on four of five conditions, Llama
on all five), so the oracle-free co-evolution of scope and operators is a property of the method, not one model.
Yield tracks DSL fluency: Gemma's expressions execute far less often (median $13\%$ vs $31\%$/$34\%$ for
Mistral/Llama), so it self-extends only on the timing seed while still never admitting a wrong operator.
\input{figures/cross_family}

\paragraph{Comparison to a text-to-SQL baseline.} As a substrate comparison (a fixed general query language vs.\ a
growing exact DSL, illustrative rather than a headline), the same model emits SQLite for each verified pair, scored
against the DSL's exact answer with the median and population-standard-deviation aggregates SQLite lacks granted.
SQL is capable but brittle, strong on flat aggregation but model-, stack-, and prompt-sensitive on ordered and
per-group queries, whereas the DSL is exact by construction, uniform across every category and model, and concise
(Table~\ref{tab:sql}, Appendix~\ref{app:sql}).

\subsection{The mechanism generalizes across domains}
\label{sec:domain-transfer}
Everything so far runs on merchant transactions. To check self-extension is a property of the mechanism, not of
one schema, we run the unmodified loop, same namespace, same gate, no new harness code beyond a schema-adapting
parser, on two further domains (Figure~\ref{fig:trajectory}). \textbf{BookSQL}~\citep{kumar2024booksql}
(business-ledger accounts): seeded from 15 question types and 9 operators, two induction rounds grow the answerable
set $15\to162$, more than tenfold, both closing a per-group gap (Appendix~\ref{app:booksql}).
\textbf{EHRSQL}~\citep{lee2022ehrsql} (clinical encounters over MIMIC-III~\citep{johnson2016mimic}): seeded from 13
types, the same loop grows the set $13\to248$, nearly twentyfold, closing a per-group gap then a conditional-filter
gap (Appendix~\ref{app:ehrsql}). Both capability families are ones the merchant operator census already spans
(Appendix~\ref{app:families}): the mechanism finds the same small set of missing capabilities wherever a
transaction-shaped schema hides them.

\subsection{Human validation of the generated items}
\label{sec:human}
As an external check, three annotators independently and blindly rated a stratified sample of $150$ generated
questions and all $53$ induced operators without seeing the gate's decisions. A majority judged $89\%$ of the DSL
expressions faithful and $97\%$ of the questions well-formed, and for all $53$ operators the annotators unanimously
agreed the two independent implementations compute the same function, the property the gate certifies automatically
(Appendix~\ref{app:human}).

%% file: figures/trajectory.tex
\begin{figure}[t]
\centering
\resizebox{\linewidth}{!}{%
\begin{tikzpicture}
\begin{groupplot}[
  group style={group size=3 by 1, horizontal sep=0.95cm},
  width=0.35\linewidth, height=3.6cm,
  xlabel={Inner-Loop Pass}, xmin=0, ymin=0,
  ymajorgrids=true, grid style={gray!18, line width=0.3pt},
  tick label style={font=\sffamily\scriptsize}, label style={font=\sffamily\footnotesize},
  title style={font=\sffamily\footnotesize, yshift=-2pt}, every axis plot/.append style={line width=0.8pt},
  legend style={font=\sffamily\scriptsize, draw=none, at={(0.5,-0.46)}, anchor=north, legend columns=2, /tikz/every even column/.append style={column sep=6pt}}]

\nextgroupplot[title={Merchant}, ylabel={Answerable Question Types}, ymax=400, xmax=24.5]
\begin{scope}[on background layer]
  \fill[gray!7] (axis cs:0,0) rectangle (axis cs:8.5,400);
  \fill[gray!7] (axis cs:16.5,0) rectangle (axis cs:24.5,400);
\end{scope}
\draw[gray!55,dashed,line width=0.5pt] (axis cs:8.5,0)--(axis cs:8.5,400);
\draw[gray!55,dashed,line width=0.5pt] (axis cs:16.5,0)--(axis cs:16.5,400);
\node[font=\sffamily\scriptsize,anchor=north,inner sep=1pt,fill=white,fill opacity=0.75,text opacity=1] at (axis cs:8.5,398) {$+6$};
\node[font=\sffamily\scriptsize,anchor=north,inner sep=1pt,fill=white,fill opacity=0.75,text opacity=1] at (axis cs:16.5,398) {$+2$};
\addplot[blue!70!black,mark=*,mark size=0.8pt] coordinates {(0,10) (1,86) (2,103) (3,116) (4,128) (5,148) (6,160) (7,173) (8,180) (9,207) (10,221) (11,235) (12,243) (13,252) (14,265) (15,274) (16,280) (17,287) (18,297) (19,303) (20,312) (21,318) (22,325) (23,330) (24,336)};
\addplot[orange!90!black,dashed,mark=square*,mark size=0.8pt] coordinates {(0,0) (8,0) (16,76) (24,116)};
\draw[densely dotted, black!50, line width=0.7pt] (axis cs:0,10) -- (axis cs:24.5,10);

\nextgroupplot[title={BookSQL}, ymax=180, xmax=24.5]
\begin{scope}[on background layer]
  \fill[gray!7] (axis cs:0,0) rectangle (axis cs:8.5,180);
  \fill[gray!7] (axis cs:16.5,0) rectangle (axis cs:24.5,180);
\end{scope}
\draw[gray!55,dashed,line width=0.5pt] (axis cs:8.5,0)--(axis cs:8.5,180);
\draw[gray!55,dashed,line width=0.5pt] (axis cs:16.5,0)--(axis cs:16.5,180);
\node[font=\sffamily\scriptsize,anchor=north,inner sep=1pt,fill=white,fill opacity=0.75,text opacity=1] at (axis cs:8.5,178) {$+4$};
\node[font=\sffamily\scriptsize,anchor=north,inner sep=1pt,fill=white,fill opacity=0.75,text opacity=1] at (axis cs:16.5,178) {$+3$};
\addplot[blue!70!black,mark=*,mark size=0.8pt] coordinates {(0,15) (1,43) (2,52) (3,57) (4,62) (5,65) (6,67) (7,71) (8,73) (9,100) (10,105) (11,119) (12,123) (13,127) (14,129) (15,132) (16,136) (17,147) (18,150) (19,152) (20,153) (21,157) (22,158) (23,159) (24,162)};
\addplot[orange!90!black,dashed,mark=square*,mark size=0.8pt] coordinates {(0,0) (8,0) (16,8) (24,23)};
\draw[densely dotted, black!50, line width=0.7pt] (axis cs:0,15) -- (axis cs:24.5,15);

\nextgroupplot[title={EHRSQL}, ymax=270, xmax=24.5]
\begin{scope}[on background layer]
  \fill[gray!7] (axis cs:0,0) rectangle (axis cs:8.5,270);
  \fill[gray!7] (axis cs:16.5,0) rectangle (axis cs:24.5,270);
\end{scope}
\draw[gray!55,dashed,line width=0.5pt] (axis cs:8.5,0)--(axis cs:8.5,270);
\draw[gray!55,dashed,line width=0.5pt] (axis cs:16.5,0)--(axis cs:16.5,270);
\node[font=\sffamily\scriptsize,anchor=north,inner sep=1pt,fill=white,fill opacity=0.75,text opacity=1] at (axis cs:8.5,267) {$+17$};
\node[font=\sffamily\scriptsize,anchor=north,inner sep=1pt,fill=white,fill opacity=0.75,text opacity=1] at (axis cs:16.5,267) {$+9$};
\addplot[blue!70!black,mark=*,mark size=0.8pt] coordinates {(0,13) (1,69) (2,83) (3,97) (4,111) (5,120) (6,124) (7,133) (8,138) (9,157) (10,166) (11,187) (12,196) (13,205) (14,209) (15,213) (16,222) (17,225) (18,227) (19,228) (20,232) (21,239) (22,244) (23,247) (24,248)};
\addplot[orange!90!black,dashed,mark=square*,mark size=0.8pt] coordinates {(0,0) (8,0) (16,48) (24,63)};
\draw[densely dotted, black!50, line width=0.7pt] (axis cs:0,13) -- (axis cs:24.5,13);
\addlegendimage{densely dotted, black!50, line width=0.7pt}
\legend{Cumulative Answerable Types, Using an Induced Operator, Initial (seed) types}
\end{groupplot}
\end{tikzpicture}%
}
\caption{One continuous closed-loop trajectory per domain. Shaded/unshaded bands are successive inner-loop
phases; each dashed vertical line is an outer-loop induction ($+n$ operators admitted); the solid line is
cumulative answerable types and the dashed orange line the subset whose expression uses an induced operator.
Two rounds grow the set $10\to336$ (Merchant, amounts/timing seed), $15\to162$ (BookSQL), and $13\to248$
(EHRSQL) under the same namespace and gate. Discussed in \S\ref{sec:results} and~\ref{sec:domain-transfer}.}
\label{fig:trajectory}
\end{figure}
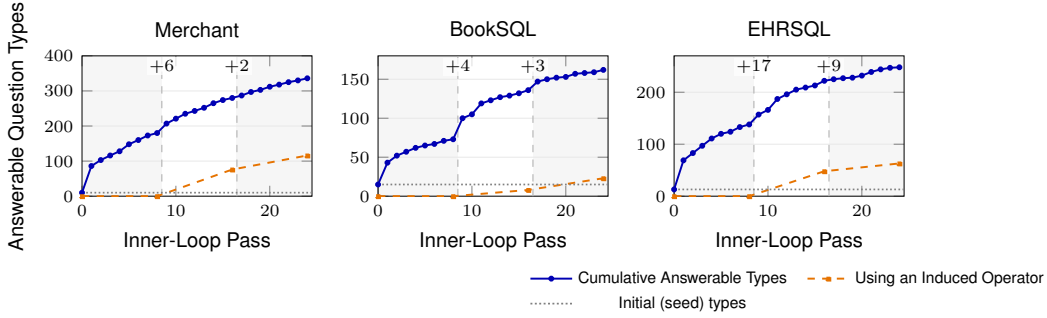

%% file: tables/gate_ablation.tex
\begin{table}[t]
\caption{Gate-component ablation on the mutation/adversarial pool (14 operators, their 138 buggy single-point
mutants, and 6 adversarial constructions; 63 merchants), each check toggled independently; the threat model is
uncorrelated (independent reference) or correlated (reference carries the same bug). \emph{Buggy}, \emph{Adv.},
and \emph{False rej.} count false-admits over the 138 mutants, admits of the 6 adversarial constructions, and
wrongful rejections of the 14 correct operators.}
\label{tab:ablation}
\centering
\small\renewcommand{\arraystretch}{0.95}
\begin{tabular}{llccc}
\toprule
Configuration & Threat model & Buggy & Adv. & False rej. \\
\midrule
Compile-only (sandbox) & n/a          & 129/138 & 6/6 & 0/14 \\
\ \ +differential      & uncorrelated & 0/138   & 4/6 & 0/14 \\
\ \ +differential      & correlated   & 129/138 & 6/6 & 0/14 \\
\ \ +metamorphic       & n/a          & 10/138  & 5/6 & 1/14 \\
Full gate              & uncorrelated & 0/138   & 4/6 & 1/14 \\
Full gate              & correlated   & 10/138  & 5/6 & 1/14 \\
\bottomrule
\end{tabular}
\end{table}

%% file: figures/base_model_ladder.tex
\begin{figure}[t]
\centering
\resizebox{\linewidth}{!}{%
\begin{tikzpicture}
\begin{groupplot}[
  group style={group size=3 by 1, horizontal sep=1.0cm},
  width=0.36\linewidth, height=3.6cm,
  ybar stacked, /pgf/bar width=10pt,
  symbolic x coords={4B,8B,14B,32B,35B}, xtick=data, enlarge x limits=0.14,
  ymin=0, ymajorgrids=true, grid style={gray!18, line width=0.3pt},
  tick label style={font=\sffamily\scriptsize}, label style={font=\sffamily\footnotesize},
  title style={font=\sffamily\footnotesize, yshift=-2pt},
  legend style={font=\sffamily\scriptsize, draw=none, at={(0.5,-0.24)}, anchor=north, legend columns=2,
    /tikz/every even column/.append style={column sep=6pt}}]

\nextgroupplot[title={Merchant}, ylabel={Answerable Question Types}, ymax=400]
\addplot[fill=blue!30, draw=blue!55!black] coordinates {(4B,10) (8B,10) (14B,10) (32B,10) (35B,10)};
\addplot[fill=blue!70!black, draw=blue!55!black] coordinates {(4B,29) (8B,47) (14B,73) (32B,225) (35B,210)};
\addplot[pattern=north east lines, pattern color=orange!90!black, draw=orange!80!black] coordinates {(4B,0) (8B,0) (14B,17) (32B,50) (35B,116)};
\node[font=\sffamily\scriptsize,anchor=south,text=orange!75!black] at (axis cs:14B,100) {+5};
\node[font=\sffamily\scriptsize,anchor=south,text=orange!75!black] at (axis cs:32B,285) {+8};
\node[font=\sffamily\scriptsize,anchor=south,text=orange!75!black] at (axis cs:35B,336) {+8};

\nextgroupplot[title={BookSQL}, ymax=200]
\addplot[fill=blue!30, draw=blue!55!black] coordinates {(4B,15) (8B,15) (14B,15) (32B,15) (35B,15)};
\addplot[fill=blue!70!black, draw=blue!55!black] coordinates {(4B,36) (8B,47) (14B,61) (32B,80) (35B,124)};
\addplot[pattern=north east lines, pattern color=orange!90!black, draw=orange!80!black] coordinates {(4B,14) (8B,9) (14B,12) (32B,49) (35B,23)};
\node[font=\sffamily\scriptsize,anchor=south,text=orange!75!black] at (axis cs:4B,65)   {+7};
\node[font=\sffamily\scriptsize,anchor=south,text=orange!75!black] at (axis cs:8B,71)   {+2};
\node[font=\sffamily\scriptsize,anchor=south,text=orange!75!black] at (axis cs:14B,88)  {+9};
\node[font=\sffamily\scriptsize,anchor=south,text=orange!75!black] at (axis cs:32B,144) {+11};
\node[font=\sffamily\scriptsize,anchor=south,text=orange!75!black] at (axis cs:35B,162) {+7};

\nextgroupplot[title={EHRSQL}, ymax=300]
\addplot[fill=blue!30, draw=blue!55!black] coordinates {(4B,13) (8B,13) (14B,13) (32B,13) (35B,13)};
\addplot[fill=blue!70!black, draw=blue!55!black] coordinates {(4B,98) (8B,67) (14B,85) (32B,117) (35B,172)};
\addplot[pattern=north east lines, pattern color=orange!90!black, draw=orange!80!black] coordinates {(4B,0) (8B,0) (14B,0) (32B,9) (35B,63)};
\node[font=\sffamily\scriptsize,anchor=south,text=orange!75!black] at (axis cs:32B,139) {+5};
\node[font=\sffamily\scriptsize,anchor=south,text=orange!75!black] at (axis cs:35B,248) {+26};
\legend{Initial (seed) types, Using a Seed Operator, Using an Induced Operator}
\end{groupplot}
\end{tikzpicture}%
}
\caption{Base-model ablation: the same outer loop run with five base models per domain, in the stacked
convention of Figure~\ref{fig:trajectory} (light: initial seed types; solid: additional types expressible with
the seed operators; hatched: types needing an induced operator; label: operators admitted). 4B/8B/14B/32B are
Qwen3 dense models and 35B is the Qwen3.6-35B-A3B main model. Read for gate soundness and whether induction
fires, not as a cross-model coverage ranking (absolute coverage is volume-sensitive; \S\ref{sec:model-ladder},
Appendix~\ref{app:model-ablation}).}
\label{fig:model-ladder}
\end{figure}
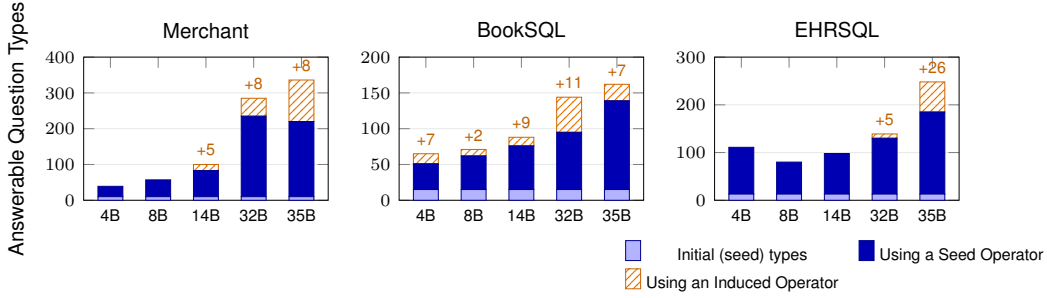

%% file: figures/cross_family.tex
\begin{figure}[t]
\centering
\resizebox{\linewidth}{!}{%
\begin{tikzpicture}
\begin{groupplot}[
  group style={group size=3 by 1, horizontal sep=1.0cm},
  width=0.36\linewidth, height=3.6cm,
  ybar stacked, /pgf/bar width=9pt,
  symbolic x coords={Qwen,Mistral,Gemma,Llama}, xtick=data, enlarge x limits=0.16,
  ymin=0, ymajorgrids=true, grid style={gray!18, line width=0.3pt},
  tick label style={font=\sffamily\scriptsize}, label style={font=\sffamily\footnotesize},
  title style={font=\sffamily\footnotesize, yshift=-2pt},
  legend style={font=\sffamily\scriptsize, draw=none, at={(0.5,-0.24)}, anchor=north, legend columns=2,
    /tikz/every even column/.append style={column sep=6pt}}]

\nextgroupplot[title={Merchant}, ylabel={Answerable Question Types}, ymax=400]
\addplot[fill=blue!30, draw=blue!55!black] coordinates {(Qwen,10) (Mistral,10) (Gemma,10) (Llama,10)};
\addplot[fill=blue!70!black, draw=blue!55!black] coordinates {(Qwen,210) (Mistral,257) (Gemma,60) (Llama,98)};
\addplot[pattern=north east lines, pattern color=orange!90!black, draw=orange!80!black] coordinates {(Qwen,116) (Mistral,35) (Gemma,0) (Llama,47)};
\node[font=\sffamily\scriptsize,anchor=south,text=orange!75!black] at (axis cs:Qwen,336)   {+8};
\node[font=\sffamily\scriptsize,anchor=south,text=orange!75!black] at (axis cs:Mistral,302) {+3};
\node[font=\sffamily\scriptsize,anchor=south,text=orange!75!black] at (axis cs:Llama,155)  {+12};

\nextgroupplot[title={BookSQL}, ymax=200]
\addplot[fill=blue!30, draw=blue!55!black] coordinates {(Qwen,15) (Mistral,15) (Gemma,15) (Llama,15)};
\addplot[fill=blue!70!black, draw=blue!55!black] coordinates {(Qwen,124) (Mistral,135) (Gemma,22) (Llama,69)};
\addplot[pattern=north east lines, pattern color=orange!90!black, draw=orange!80!black] coordinates {(Qwen,23) (Mistral,8) (Gemma,0) (Llama,25)};
\node[font=\sffamily\scriptsize,anchor=south,text=orange!75!black] at (axis cs:Qwen,162)   {+7};
\node[font=\sffamily\scriptsize,anchor=south,text=orange!75!black] at (axis cs:Mistral,158) {+1};
\node[font=\sffamily\scriptsize,anchor=south,text=orange!75!black] at (axis cs:Llama,109)  {+2};

\nextgroupplot[title={EHRSQL}, ymax=300]
\addplot[fill=blue!30, draw=blue!55!black] coordinates {(Qwen,13) (Mistral,13) (Gemma,13) (Llama,13)};
\addplot[fill=blue!70!black, draw=blue!55!black] coordinates {(Qwen,172) (Mistral,134) (Gemma,62) (Llama,81)};
\addplot[pattern=north east lines, pattern color=orange!90!black, draw=orange!80!black] coordinates {(Qwen,63) (Mistral,0) (Gemma,0) (Llama,12)};
\node[font=\sffamily\scriptsize,anchor=south,text=orange!75!black] at (axis cs:Qwen,248)   {+26};
\node[font=\sffamily\scriptsize,anchor=south,text=orange!75!black] at (axis cs:Llama,106)  {+5};
\legend{Initial (seed) types, Using a Seed Operator, Using an Induced Operator}
\end{groupplot}
\end{tikzpicture}%
}
\caption{Cross-family transfer: the same closed loop, same frozen prompt and gate, with four proposer
\emph{families} (Qwen3.6-35B-A3B, Mistral-Small-24B, Gemma-3-27B, Llama-3.3-70B) on the three domains, in the
stacked convention of Figure~\ref{fig:model-ladder}. The gate stays sound for every family (zero false admits
throughout), and two further families self-extend (Mistral on Merchant and BookSQL, Llama on all three). Read for
soundness and induction, not coverage ranking; per-family compile rates and the three-seed companion are in
Table~\ref{tab:cross-family} (\S\ref{sec:model-ladder}).}
\label{fig:cross-family}
\end{figure}
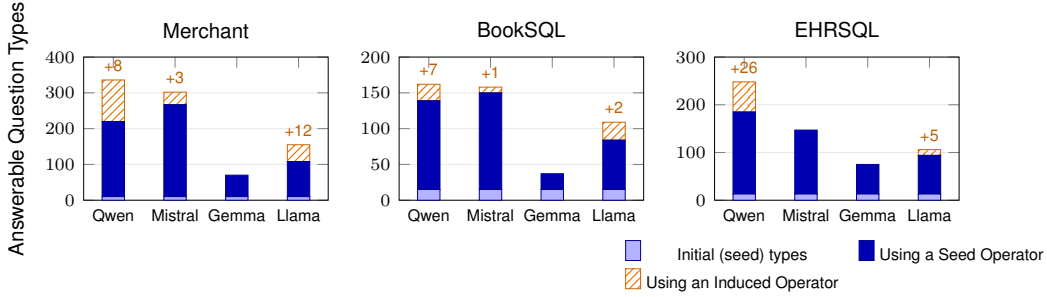

%% file: sections/related_work.tex
\section{Related work}
\textbf{Library learning / DSL induction.} DreamCoder \citep{ellis2021dreamcoder} learns reusable abstractions by
wake-sleep library learning, LILO \citep{grand2024lilo} compresses LLM-written programs into a reusable library, and
TroVE \citep{wang2024trove} grows and trims a toolbox of functions; all build reusable abstractions from solved or
executable tasks, whereas our gate uses no answer key, admitting only on agreement between two independent
implementations and conservation laws over trusted operators.
We isolate this: the LLM-judge baseline (Appendix~\ref{app:judge}) and the compile-only ablation
(Table~\ref{tab:ablation}) are the ``admit by judgment'' and ``admit if it runs'' proxies, both admitting wrong
operators the gate rejects. \textbf{Verifier-guided program discovery.} FunSearch \citep{romera2024mathematical}
and AlphaEvolve \citep{novikov2025alphaevolve} score candidates against an objective and keep the best; our gate has
no objective and no competition, admitting or rejecting once by agreement and conservation.
\textbf{Self-extending agents.} Voyager \citep{wang2024voyager}, LATM \citep{cai2024large}, and CREATOR
\citep{qian2023creator} grow skill/tool libraries where a tool earns its place by working on tasks or by the
model's judgment, whereas ours are exact DSL operators admitted with no task reward and no judge.
\textbf{Program-of-thoughts.} PoT \citep{chen2023pot} has the model emit a program and delegates the computation to
an interpreter over a fixed set of primitives; our agent is likewise a program emitter, but its operator set grows
through oracle-free induction, so PoT is the special case of one fixed operator set. \textbf{Verification without a judge.} Our checks combine hardened
self-consistency \citep{wang2023selfconsistency}, differential testing \citep{mckeeman1998differential,yang2011csmith},
and oracle-free metamorphic testing \citep{chen1998metamorphic,segura2016survey} into an admission gate for
self-authored operators; like POET \citep{wang2019paired}, our asker co-evolves against the operator library.

%% file: sections/conclusion.tex
\section{Conclusion and limitations}
We presented a two-loop, single-model system that grows its own DSL under an oracle-free admission rule: an inner
loop triples coverage, and an outer loop invents operators verified with no answer key. It transfers unmodified to
two further domains and stays sound across four model families.

\paragraph{Limitations.}\label{sec:mode-limitation}
(i) The most-requested capability across every seed and domain, a most-frequent-value (mode) reducer, is
permanently out of reach: it is the top-ranked gap everywhere and has no conservation law, so only the mode
\emph{value} (a non-numeric argmax) stays outside, while per-group count and frequency are admitted like per-group
sum (Appendix~\ref{app:reach}). (ii) We demonstrate up to three induction rounds per seed across all three seeds,
none stalling (Appendix~\ref{app:ops}); scaling to many rounds is untested. (iii) Admission yield is modest, most
drafts rejected because the different-style reference reverts to sandbox-forbidden method calls ($81\%$ of
rejections), a throughput cost that leaves soundness unaffected (Appendix~\ref{app:false-reject}). (iv) A declared
relation only binds if it names another trusted operator (Section~\ref{sec:adversarial}): a relabeling defeats the
gate harmlessly (no legal question surfaces it), and a size-changing bug escapes if its only witnessing relation is
a bare builtin, needing a sandbox-level fix. (v) All three seeds center on per-group aggregation, what transaction
QA most demands; the 16 operators still span four families (Appendix~\ref{app:families}), but a seed around a
qualitatively unrelated gap is untested, though we test transfer across \emph{domains} (\S\ref{sec:domain-transfer}).

%% file: sections/repro_statement.tex
\section*{Reproducibility Statement}
All merchant-transaction experiments run on a fully synthetic corpus of merchant transaction histories
(Appendix~\ref{app:dataset}) that matches the distribution of real merchant transactions without using any
customer data; the supplementary code includes the three generator scripts and the aggregate parameter file they
sample from, so the corpus regenerates by a single command with no access to proprietary data. The two
domain-transfer studies (\S\ref{sec:domain-transfer}) instead run on BookSQL and MIMIC-III/EHRSQL, external
research benchmarks obtained from their own sources under their own release terms
(Appendices~\ref{app:booksql},~\ref{app:ehrsql}). The supplementary code contains the DSL and its tests, the
inner- and outer-loop agent with the exact prompts, every study and ablation, the seed question sets, and a
top-level \texttt{README} with the environment setup and per-script commands. Given the sampled expressions,
every admission check (execution, value-signature comparison, differential and metamorphic tests) is
deterministic, so every reported result regenerates from the released scripts.

%% file: sections/ethics_statement.tex
\section*{Ethics Statement}
All merchant-transaction experiments (everything outside \S\ref{sec:domain-transfer}) run on a fully synthetic
corpus (Appendix~\ref{app:dataset}) containing no real customer data. The domain-transfer study additionally uses
two existing public research benchmarks, rather than any newly collected data: BookSQL (business-ledger records)
and EHRSQL (clinical encounters). EHRSQL is built over MIMIC-III, a de-identified critical-care database released
for research under PhysioNet's credentialed-access Data Use Agreement; we obtained MIMIC-III through
PhysioNet's credentialed-access process and reproduced our EHRSQL results from it. We
attempt no re-identification, and this paper reports only aggregate, cohort-level statistics (accuracy
percentages, operator counts, coverage numbers); no patient-level record, free-text note, or identifier appears
anywhere in the paper.

%% file: sections/llm_usage_statement.tex
\section*{Use of Large Language Models}
In preparing this manuscript, the authors used an AI coding assistant (Claude Code) to draft and revise prose in
the results and appendix sections, and to produce and iterate on LaTeX figures and tables.

%% file: sections/dataset.tex
\section{Dataset: a synthetic transaction corpus}
\label{app:dataset}
Every merchant-transaction experiment in this paper (all results outside \S\ref{sec:domain-transfer}) runs on a
\emph{fully synthetic} corpus of merchant transaction histories.
\textbf{No real user or customer data is used anywhere in these experiments}: not in training, validation, induction,
gate evaluation, or the held-out check. (The domain-transfer studies use BookSQL, a public benchmark, and
EHRSQL over MIMIC-III, a credentialed-access clinical database under PhysioNet's Data Use Agreement; both are
unrelated to this corpus, and Appendices~\ref{app:booksql},~\ref{app:ehrsql} describe their own, separate
provenance.) To keep the corpus realistic, we fit \emph{only aggregate} statistics
of a real merchant-transaction distribution (quantile grids and frequency tables, never individual records) and
sample fresh merchants from those statistics under standard generative assumptions (empirical marginals, a Zipf
buyer-frequency skew, and clustered active-day timestamps). The synthetic corpus therefore matches the real
distribution on the properties the operators and questions exercise, while no record, identifier, timestamp, or
field is copied from or derived from any individual. The aggregate-fitted generator and the aggregate parameter
file it samples from are free of customer-derived rows, so the corpus regenerates, and the entire pipeline runs,
with no access to any proprietary data.

\paragraph{What a record is.} Each merchant is a chronological sequence of transactions; each transaction has an
amount, a timestamp (with weekday, hour, date), a buyer identifier, a buyer country, a buyer account age, and a
guest/registered flag, plus per-merchant scalars (category, country, account age, distinct-buyer and
transaction counts). This is the schema the interpreter parses (Appendix~\ref{app:prompts}).

\paragraph{Generation (three scripts, all in the supplementary code).}
\texttt{synth\_stats.py} reads a real corpus and writes \emph{only} aggregate statistics to a params file:
quantile grids for amounts, transaction counts, account ages, inter-transaction gaps, and the distinct-buyer and
active-day ratios; frequency tables for weekday, hour, buyer country, category, and merchant country; a guest
rate; and a Zipf exponent for buyer-frequency skew. No raw record leaves this step. \texttt{synth\_gen.py} then
samples fresh merchants: transaction counts, amounts, and ages from the empirical marginals; timestamps by
placing transactions on a smaller set of weekday-seasonal \emph{active days} (reproducing the bimodal
intra-day/inter-day gap structure); buyers by a Zipf process with guaranteed coverage (reproducing the
buyer-frequency skew per-group operators rely on); and \emph{freshly minted} buyer identifiers. Per-buyer
country, guest status, and account age are held consistent across a buyer's transactions.

\paragraph{Fidelity.} We validate the match by re-extracting the aggregate statistics from the synthetic corpus
and comparing to the real params (\texttt{synth\_fidelity.py}). On the operator-relevant fields the match is
tight: median relative error is $0.000$ for transaction count, $0.002$ for merchant age, $0.014$ for the
distinct-buyer ratio, $0.004$ for amount, and $0.054$ for buyer age; the total-variation distance between the
real and synthetic frequency tables is $\le 0.02$ for weekday, hour, buyer country, category, and merchant
country; the guest rate ($0.090$ vs $0.090$) and buyer-frequency Zipf exponent ($0.90$ vs $0.92$) match. The
inter-transaction gap distribution matches in scale and tail (median $5.0$h vs $7.2$h, $90$th percentile $324$h
vs $309$h); it slightly under-represents the near-instant burst tail, which does not affect the results, since
the gap-based operators test exact computation on whatever gaps occur.

\paragraph{Sizes.} We generate $12{,}000$ synthetic merchants; the coverage, induction, mutation, and
adversarial studies each draw a 63-merchant validation set from this corpus. The DSL's expression map is
additionally cross-checked against the independent gold extractors, whose comparison logic is validated by a
self-test over all 50 question types (agreement is a property of the two deterministic implementations, not of
the corpus).

%% file: sections/appendix.tex
\section{Algorithm listing}
\label{app:algo}
Figure~\ref{fig:loops} (main text, Section~\ref{sec:method}) diagrams both loops; Algorithm~\ref{alg:selfextend}
gives them as pseudocode.

\input{algo/self_extend}

\section{Reproducibility and environment}
\label{app:repro}
All runs use a single frozen base model, \texttt{Qwen3.6-35B-A3B}, served with vLLM on NVIDIA B200 GPUs (one GPU
per job). The model is never fine-tuned and its reasoning traces are disabled; sampling uses temperature $1.0$
and top-$p$ $0.95$. Randomness enters only through sampling: given the sampled expressions, every downstream
check (execution, value-signature comparison, differential and metamorphic tests) is deterministic. The default
budgets are: $64$ operator drafts per induction, $24$ generations for autonomous gap discovery with $12$
questions requested each, $1$ repair retry, $4$ independent answers per question in verification, and up to
$2048$ new tokens per generation. The outer-loop induction, mutation soundness, adversarial-operator,
independence, repeated-run variance, admission-baseline, operator-family, and text-to-SQL studies are each
regenerated by a single fixed command; the released \texttt{inner\_loop/OUTPUTS.md} maps every result file each
command produces to the table, figure, or number it backs. Table~\ref{tab:closed}, for instance, is produced by
the \texttt{closed\_loop\_ps\_*} runs (63 merchants), not the earlier same-named 203-merchant
\texttt{closed\_loop\_\{round1,buyer,timing\}} run that survives only as the text-to-SQL script's default
evaluation set. The exact commands, the harness, prompts, and seed sets are provided in the supplementary
materials.

\section{The DSL sandbox and the method-free operator subset}
\label{app:sandbox}
Two sandboxes enforce the exact-verifiability invariant.

\paragraph{Expression sandbox (answers).} A candidate answer is a single expression, parsed to an abstract
syntax tree and evaluated under an allowlist of node types. It permits literals and collections, arithmetic and
comparison, indexing and slicing, and calls whose target is a bare registered name. It forbids attribute access
(no \texttt{x.y}), imports, the \texttt{**} operator, comprehensions with side effects, assignment, any name
beginning with a double underscore, and calls to arbitrary expressions. Built-ins are emptied and only the
registered operators are in scope. A per-evaluation wall-clock timeout (via \texttt{SIGALRM}) bounds every call.

\paragraph{Operator-body sandbox (induced operators).} A new operator is one method-free function definition,
compiled under a superset of the expression allowlist that additionally permits \texttt{def}, arguments,
\texttt{return}, assignment and augmented assignment, bounded \texttt{for}, \texttt{if}, \texttt{pass},
\texttt{break}, \texttt{continue}, and list/dict/set comprehensions. It still forbids attribute access and
therefore any method call (no \texttt{.append}, \texttt{.get}, \texttt{.items}), imports, \texttt{while},
\texttt{**}, dunder names, and recursion. A call target must be a bare name or an element of a local table,
\texttt{tbl[key](...)}, which supports a dispatch table of reducers without a method call. The available
callables are a fixed builtin whitelist (\texttt{len, sum, min, max, sorted, zip, range, enumerate, set, dict,
list, any, all, abs, round}) plus the operators already trusted. Each induced operator runs under the same
per-call timeout. Because the subset is first-order and attribute-free, an induced operator cannot escape to
arbitrary computation, so the growing library stays auditable with no supervision.

\section{Prompts}
\label{app:prompts}
All prompts are issued to the same frozen model. The data schema shared by the prompts is:
\begin{lstlisting}
Per-transaction lists (index-aligned): amounts ($ floats), timestamps (datetimes),
hours (0-23), weekdays (names), dates, buyer_ids, buyer_countries,
buyer_ages_days, is_guest.
Merchant scalars: total_transactions, distinct_buyers, merchant_age_days,
merchant_category, merchant_country.
\end{lstlisting}

\paragraph{Autonomous gap discovery.} The model is given only the current operators and asked to propose natural
questions, writing a new operator name where the current set cannot express one:
\begin{lstlisting}
You are a financial-data analyst with a small calculator over one merchant's
transactions. You have ONLY these operators: <library>. Data columns: <schema>.
Propose diverse, natural analytics questions a merchant would actually ask. For
each, write the single expression you would use to answer it. If the operators
above CANNOT express it, still write the expression using a NEW operator you wish
existed, giving it a clear descriptive name and plausible arguments -- do not
force-fit the existing operators. This is how we discover what is missing.
\end{lstlisting}
An expression that calls a name outside the current library marks a gap; the loop tallies these and induces the
most-requested capability.

\paragraph{Operator induction.} The authoring prompt states the method-free rules and a strict output format
(name, signature, spec, implementation, an independently written reference, metamorphic relations, and example
calls). The rules are:
\begin{lstlisting}
A new operator is a Python function run in a SANDBOX that allows ONLY a
METHOD-FREE subset: def, return, assignment, for, if, and comprehensions;
builtins len,sum,min,max,sorted,zip,range,enumerate,set,dict,list,any,all,abs,
round; the existing DSL operators -- and NOTHING else (no group-by, no windowing;
inventing one is the whole point). STRICTLY FORBIDDEN (function rejected):
attribute access or ANY method call (.append,.get,.items,...); imports; while;
**; dunder names; recursion. A local dispatch table IS allowed, e.g.
reducers={'sum':sum,'max':max} then reducers[op](xs). Group with a comprehension
and reduce each group with a listed reducer; RETURN A LIST (one number per group).
Build lists with a comprehension or "out = out + [x]", NEVER "out.append(x)". The
REFERENCE must also be method-free, written in a different style.
\end{lstlisting}

\paragraph{Text-to-SQL baseline.} For the baseline the same model is asked for one SQLite query:
\begin{lstlisting}
You answer analytics questions over one merchant's transactions using SQLite.
Table `txns`, one row per transaction, columns: amount REAL, timestamp REAL (unix
seconds), hour INTEGER, weekday TEXT, date TEXT, buyer_country TEXT,
buyer_age_days INTEGER, is_guest INTEGER, buyer_id TEXT.
Two extra aggregate functions are available, matching a full analytics SQL engine:
median(x) and stddev(x) (population standard deviation).
Write ONE SQLite query that returns the answer as a single numeric value (one row,
one column). Output ONLY the SQL, nothing else.
\end{lstlisting}

\section{The three starting sets}
\label{app:seeds}
A starting set is a small collection of question types together with the minimal operator set that answers them,
and no others; an induced operator is genuinely new only if it lies outside that set's closure, so we never leak
a target capability into the starting library. We run the closed outer loop on three independent starting sets,
each of which \emph{induces from its own} minimal operators (the gap-discovery and induction prompts list exactly
that set's operators):
\begin{itemize}
\item \emph{amounts/timing}: \texttt{mean, min, max, median, std, sum, ratio, gaps\_hours, gaps\_days} (9);
\item \emph{top-buyer}: \texttt{max, mean, median, nunique, pct\_of, ratio, sum, top\_count} (8);
\item \emph{timing}: \texttt{days\_between, earliest, first\_half, gaps\_days, gaps\_hours, int, max, mean,
median, nunique, percent\_change, ratio, second\_half, std, sum} (15).
\end{itemize}
None of the three can express per-group aggregation, and in each the loop discovers a gap it cannot express (a
mode reducer for amounts/timing and timing, a conditional-proportion reducer for top-buyer) and induces a
per-group operator
demonstrably outside that set's closure, then re-runs the inner loop on that set's question types.
The outer-loop result is thus genuinely across three independent starting sets. The inner-loop robustness sweep
is separate: it varies the operator set across fourteen starting sets, each yielding a strictly positive number
of new verified types.

\input{figures/seed_trajectory}

Table~\ref{tab:saturation} gives the full round-by-round funnel, proposed, executable, verified, behind the
coverage numbers cited in Section~\ref{sec:inner}.

\input{tables/saturation}
\input{tables/closed_loop}

\section{The induced operators}
\label{app:ops}
\texttt{per\_group\_sum} is a representative admitted operator, authored method-free from the seed
reducers. Its implementation and the independently written reference it must agree with, both emitted method-free in
the induction response and shown here verbatim from a run, are:
\begin{lstlisting}
def per_group_sum(keys, values):             # implementation
    filtered = [(k, v) for k, v in zip(keys, values) if k is not None and v is not None]
    if not filtered:
        return []
    unique_keys = sorted(set(k for k, _ in filtered))
    out = []
    for k in unique_keys:
        group_vals = [v for kk, v in filtered if kk == k]
        out = out + [sum(group_vals)]
    return out

def group_addition(keys, values):            # independent reference
    pairs = [(x, y) for x, y in zip(keys, values) if x is not None and y is not None]
    if not pairs:
        return []
    distinct = sorted(set(p[0] for p in pairs))
    result = []
    for key in distinct:
        total = 0
        for pair in pairs:
            if pair[0] == key:
                total = total + pair[1]
        result = result + [total]
    return result
\end{lstlisting}
Both drop \texttt{None} keys and values, group by the sorted unique key, and sum each group, so they agree
element-wise despite using different filtering styles (a list comprehension vs.\ an explicit tuple-pair loop).
Differential testing compares their return values on every validation
merchant and on fuzzed namespaces, empty, singleton, duplicate-key, and \texttt{None}-key, each also with the
array order reversed, here $0$ disagreements over $213$ checks. The comparison is order-sensitive, so the
reversed-order fuzz is precisely what forces the two implementations to commit to the same deterministic
grouping order. Two of the metamorphic relations the model declared, each tying the operator to a trusted
primitive, are:
\begin{lstlisting}
sum(per_group_sum(buyer_ids, amounts)) == sum(amounts)
sum(per_group_sum(buyer_countries, amounts)) == sum(amounts)
\end{lstlisting}
The reference need not resemble the implementation. An admitted \texttt{per\_group\_max} pairs a
comprehension-style implementation with an explicit dictionary-and-loop reference, and the two must still agree
on every input ($0/213$):
\begin{lstlisting}
def per_group_max(keys, values):             # implementation
    uniq = sorted(set(k for k in keys if k is not None))
    return [max([v for kk, v in zip(keys, values) if kk == k]) for k in uniq]

def per_group_max_ref(keys, values):         # independent reference
    groups = {}
    for k, v in zip(keys, values):
        if k not in groups:
            groups[k] = []
        groups[k] = groups[k] + [v]
    uniq = sorted(set(k for k in keys if k is not None))
    result = []
    for k in uniq:
        vals = groups[k]
        best = None
        for x in vals:
            if best is None or x > best:
                best = x
        result = result + [best]
    return result
\end{lstlisting}
Across the three starting sets the gate admitted 6, 5, and 2 distinct operators respectively, almost all
per-group aggregations outside that set's closure (amounts/timing's admitted set also includes one
whole-corpus utility, \texttt{distinct\_count}, needed to express a seed-adjacent question but not itself
grouped). The amounts/timing set admitted \texttt{per\_group\_max}, \texttt{group\_median},
\texttt{grouped\_sum}, \texttt{distinct\_count}, \texttt{per\_group\_sum}, and \texttt{per\_group\_total}; the
top-buyer set admitted \texttt{per\_flag\_split\_ratio}, \texttt{pct\_by\_flag}, \texttt{pct\_of\_boolean},
\texttt{group\_count\_ratio}, and \texttt{pct\_group}; the timing set admitted
\texttt{group\_first\_occurrence\_index} and \texttt{group\_agg}. The most-requested capabilities
were a mode reducer (amounts/timing and timing) and a conditional-proportion reducer, closing a
\texttt{where}-style gap (top-buyer). Per-group count and frequency
operators \emph{are} admitted, on the conservation law that group counts sum to the number of records
(e.g.\ \texttt{sum(per\_group\_count(buyer\_ids)) == total\_transactions}), so the mode's frequency is reachable
as \texttt{max} of such a vector; only the mode value itself, an argmax over keys, has no conservation law and
is not admitted.

\paragraph{A third round.} Table~\ref{tab:closed}'s round-2 row already chains one induction onto another via
\texttt{--prior-ops}, loading a prior round's admitted operators into the library before inducing again. We chain
a third round the same way onto all three seeds' first two rounds. On amounts/timing (18 operators in the
library), the autonomously discovered gap is a \texttt{new\_filter}-style capability, and the gate admits an
operator, \texttt{group\_and\_sum\_keys}, on the first attempt, no fallback gap needed, newly verifying 56 more
question types. On top-buyer, the discovered gap is a uniqueness-style capability and the gate admits three
operators (\texttt{pct\_by\_country}, \texttt{group\_pct\_share}, \texttt{pct\_by\_category}), newly verifying
154 more types. On timing, the top-ranked gap is again \texttt{mode}, and the gate admits two operators
(\texttt{per\_day\_counts}, \texttt{group\_agg\_sum\_by\_keys}), newly verifying 207 more types. All three seeds
self-extend into a third round; none stalls. This is still not a systematic sweep of many rounds, only evidence
the loop does not stop at two, not a claim that three is a general ceiling.

\section{Anatomy of an induction: authored, unlocked, and rejected}
\label{app:anatomy}
We walk through one induction end to end on the amounts/timing set, from the committed run of
Figure~\ref{fig:seed-trajectory}. The autonomously discovered gap is a \emph{mode} reducer, surfaced by questions such
as ``what is the most frequent buyer country?'', ``the most common hour of the day?'', and ``the most frequent
weekday?''. The mode \emph{value} has no conservation law and is not admitted; the model instead authors the
per-group operators around it that \emph{do} (Appendix~\ref{app:ops}), and the loop admits six this round
(among them \texttt{group\_agg\_by}, \texttt{count\_per\_group}, \texttt{group\_sum}, \texttt{group\_max}, and
\texttt{per\_group\_total}).

\paragraph{What it unlocks.} With these in the library the inner loop verifies question types no composition of
the base operators could express, for example:
\begin{lstlisting}
Highest transaction count in a single hour of day? -> max(count_per_group(hours, amounts))
The merchant's best day by total sales volume?     -> max(group_sum(dates, amounts))
Std. dev. of the transaction count per buyer?      -> std(count_per_group(buyer_ids, amounts))
\end{lstlisting}
The \emph{count} of the mode is thus reachable (\texttt{max} of a per-group count) even though the mode value is not.

\paragraph{What is rejected, and why.} The gate turns drafts away on the two failure modes it should. Most are
rejected by the \emph{substrate}: a \texttt{mode} draft that reaches for method calls (\texttt{counts.values()},
\texttt{counts.items()}) or a \texttt{lambda} never compiles under the method-free sandbox. A draft that
\emph{compiles} can still be rejected on \emph{semantics}. The model proposed a correct per-group mean but
attached a false conservation claim:
\begin{lstlisting}
def group_mean(keys, values):
    groups = {}; order = []
    for k, v in zip(keys, values):
        if v is None: continue
        if k in groups: groups[k] = groups[k] + [v]
        else: groups[k] = [v]; order = order + [k]
    results = []
    for k in order:
        results = results + [sum(groups[k]) / len(groups[k])]
    return results
# declared relation (FALSE): sum(group_mean(buyer_ids, amounts)) == sum(amounts)
\end{lstlisting}
The implementation is a valid per-group mean, but the metamorphic relation the model declared for it is false:
per-group means do not sum to the global total. The gate evaluates that relation on the data and rejects the
draft. Declared invariants are checked, not taken on faith: a plausible draft whose stated invariant does not
hold is turned away (and this is one reason admission yield is modest: a valid operator can be lost when the
model attaches a bad relation to it).

\section{Oracle-free validation gate}
\label{app:gate}
An operator is admitted only if all of the following hold. \textbf{(1) Differential.} Its two implementations
return the same value on every declared call, over the synthetic merchants and over fuzzed inputs: empty, singleton,
duplicate-key, containing \texttt{None}, and each of these with the record order reversed. \textbf{(2)
Metamorphic.} Every declared relation holds on every synthetic merchant, and at least one relation is non-trivial: it
must call the new operator \emph{and} a trusted operator and be a comparison, so a relation that is just
\texttt{True} is rejected; equalities are compared with a small floating-point tolerance, checked on the
validation records rather than the fuzz battery differential testing above uses, a conservation law being a
claim about what the data contains, not about synthetic edge cases. \textbf{(3) Round-trip.} After
admission, every new question the operator unlocks must still pass the inner loop's independent-agreement check.
No single test is an answer key; a wrong operator would have to pass all of them together.

\section{Verified compression}
\label{app:compress}
Compression mines recurring structure from the verified expressions the loops produced and folds it into new
named operators, admitted by equivalence rather than by task reward. From the 219 verified expressions we abstract
each recurring sub-expression's data-column and constant leaves into parameters (a leaf shared across occurrences
becomes a shared parameter), rank the resulting templates by reuse, and for each candidate synthesize a
method-free operator whose body is the template. The gate admits it only if it compiles under the operator-body
sandbox (Appendix~\ref{app:sandbox}) and every occurrence agrees with the abstraction on all 63 merchants, an
oracle-free equivalence check that needs no answer key. Table~\ref{tab:compress} lists the five most-reused
patterns, all admitted this way; the ``checks'' column is the number of (occurrence $\times$ merchant) equivalence
evaluations each abstraction had to pass.

\input{tables/compress}

\section{Mutation study details}
\label{app:mutation}
For each admitted operator we generate every single-point mutant of its implementation with a fixed set of
mutation operators: swap an arithmetic operator ($+\leftrightarrow-$, $*\leftrightarrow/$), swap a comparison
(including \texttt{is}/\texttt{is not}), perturb a numeric constant by $\pm 1$, swap a called reducer for another
(\texttt{sum} to \texttt{max}, and so on), swap the first two arguments of a call, swap the two function
parameters, and negate an \texttt{if} guard. A mutant that matches the original on every input, evaluated over
the synthetic merchants plus the fuzz battery, is an equivalent mutant and is discarded; the original operator is used
only for this labeling and the gate never sees it. Each remaining buggy mutant is run through the gate under two
threat models: \emph{uncorrelated}, in which only the implementation is mutated and the reference stays correct,
and \emph{correlated}, in which the same mutation is written into both implementations, so the differential check
cannot catch it; only the metamorphic and structural checks can. Table~\ref{tab:mutation}
reports the per-operator outcome; the gate's false-admit rate is $0/138$ under an independent reference, and
$10/138$ under the correlated worst-case (the identical bug written into both implementations, defeating the
differential check, the regime independence is designed to prevent).

\input{tables/mutation_soundness}

Re-running the full gate on the same candidate pool at $N=3,10,30,63$ merchants leaves every rate flat: the
buggy false-admit rate stays $0/138$, the correct-operator false-reject rate stays $1/14$ (the same isolated miss
as Table~\ref{tab:ablation}), and the one correlated-threat adversarial weak point (Section~\ref{sec:adversarial})
stays $5/6$; the gate's outcome is invariant to the validation-set size.

\section{Adversarial operator details}
\label{app:adversarial}
\texttt{group\_size} (signature \texttt{group\_size(keys)}, a per-key count) is admitted from the timing
starting set on:
\begin{lstlisting}
sum(group_size(buyer_ids)) == total_transactions
sum(group_size(hours)) == total_transactions
mean(group_size(buyer_ids)) == ratio(total_transactions, distinct_buyers)
\end{lstlisting}
Together with \texttt{per\_group\_sum} (Appendix~\ref{app:ops}), we hand-write three wrong implementations of
each: \emph{cyclic} reports each key's slot using the previous key's true value (a rotation by one),
\emph{swap} transposes exactly the first two keys' true values, and \emph{merge} adds the first two keys' true
values into one slot and drops the other. Cyclic and swap permute the same multiset of per-key values the
correct operator produces, so \texttt{sum}, \texttt{len}, \texttt{min}, \texttt{max}, and \texttt{mean} of their
output equal the correct operator's on every input; merge is not a permutation and serves as a control. Each
of the six mutants is run through \texttt{validate\_operator} (Appendix~\ref{app:gate}) against the same
63-merchant validation set the mutation study uses (63 synthetic merchants; Appendix~\ref{app:dataset}),
under the uncorrelated and correlated threat models of
Appendix~\ref{app:mutation}.
Table~\ref{tab:adversarial} reports every cell; the constructions are deterministic and the outcomes follow from
the relations and the constructions, not from properties specific to this sample.

The one false admit, \texttt{per\_group\_sum} + merge under the correlated model, depends on which of its four
relations (Appendix~\ref{app:ops}) are size-sensitive. \texttt{len(per\_group\_sum(buyer\_ids, amounts))
== distinct\_buyers} would catch a merge, since dropping a group shortens the output list, but non-triviality
(Appendix~\ref{app:gate}) requires a relation to name a trusted \emph{operator}, and \texttt{len} is a builtin
available inside an operator's own body (Appendix~\ref{app:sandbox}) but is not itself a registered operator, so
this relation never qualifies and is never evaluated. Widening non-triviality to accept it would not be enough
on its own: relations execute in the same restricted namespace as answer expressions, which resolves only
registered operator names and never raw builtins, so the relation would still fail to evaluate.
Closing this needs relations to reach the same builtin surface an operator's body already can, a sandbox-level
change, not a filter fix; we leave it to future work. \texttt{per\_group\_sum}'s other three relations are
size-insensitive sums and a lower bound, so none can distinguish a merge; \texttt{group\_size}'s single
size-sensitive relation is written with \texttt{mean} and \texttt{ratio}, both registered operators, so
it is checkable and correctly kills the same construction.

Finally, we test whether relabeling is exploitable downstream, i.e.\ whether any legal answer expression could
look up one specific key's value from a per-group result rather than only aggregate over it. We evaluate
\texttt{zip([1], [2])}, \texttt{sorted([2, 1])}, and \texttt{set([1, 1])} directly in the same expression
sandbox a question's answer runs in; all three fail with a name error, confirming Appendix~\ref{app:sandbox}'s
restriction empirically rather than only by reading it. No answer expression can build the keyed lookup a
relabeling would need to be exposed downstream.

A natural follow-up is whether this protection survives if the DSL later grows to expose such a lookup. We
re-validate each relabeling with one added differential call that reads the per-group list directly, not through
a symmetric aggregate, simulating exactly that growth. Under an independent reference this rejects every
relabeling ($4/4$: \texttt{cyclic} and \texttt{swap} for both \texttt{per\_group\_sum} and \texttt{group\_size}),
flipping them from admitted to rejected, while the correct operators still pass. Under the correlated model,
where the same buggy source serves as its own reference, the added call leaves all four still admitted: two
copies of identical code cannot disagree on any call one adds, the same limit differential testing meets against
correlated bugs throughout this paper, and here the metamorphic layer cannot back it up either, since every
declared relation is a symmetric aggregate, inherently blind to relabeling. So an admitted operator's
soundness is relative to the downstream surface available at admission time: extending that surface and
re-validating catches a previously-harmless error, but only when the re-validation reference is independent of
the original bug.

\input{tables/adversarial}

\section{LLM-judge admission baseline}
\label{app:judge}
A natural alternative to the oracle-free gate is to let the same frozen model judge each operator's correctness
directly. We ask it exactly that over the same pool as the soundness studies (the 14 admitted operators, their 138
buggy single-point mutants, and the 6 adversarial constructions), majority vote over 5 samples per candidate with
reasoning enabled; only 11 of 790 generations produced no parseable verdict. Table~\ref{tab:judge} reports the
result against the gate. The judge is not unsound on this pool: it admits none of the 138 buggy mutants, and its
blanket suspicion even rejects all 6 adversarial relabelings the gate admits. But it is unusable as an admission
rule. First, it is imprecise: it wrongly rejects 2 of the 14 correct operators. In each case the operator's own
free-text specification, authored alongside it during induction, does not quite match its own implementation: a
per-key sum specified to preserve first-appearance order but implemented sorted by key, and a per-key max
specified as ``paired with their corresponding values'' but implemented as a bare list of maxima. These mismatches
are harmless to the gate, since admission never checks specification text against code, only differential
agreement and conservation; the judge, reading the specification literally, treats each as disqualifying. A loop
gated this way would discard half the valid capabilities it induces. Its apparent soundness is the same literal
strictness, not principled verification: it rejects whatever the prose does not exactly match, which happens to
include the relabelings and, equally, correct operators. Second, and more fundamentally, a judge's verdict is a
probabilistic opinion with no checkable basis, whereas the gate's admission is a deterministic, re-runnable
check, differential agreement plus conservation against trusted primitives, that an auditor can reproduce. On
this pool the gate is both sound and more precise than the judge: it wrongly rejects only 1 of the 14 correct
operators where the judge rejects 2, and its verdicts are reproducible where the judge's are not. The 4 adversarial
relabelings the gate admits are observationally equivalent under the current DSL (Appendix~\ref{app:adversarial}),
wrong to no expressible question.

\input{tables/judge_baseline}

\section{Text-to-SQL baseline details}
\label{app:sql}
The evaluation set is the verified (question, expression) pairs the loops produced; the DSL expression is the
exact gold. Each question is categorized by the operators its expression uses: \emph{flat} (whole-table
reduction), \emph{per-group} (a per-key reduction, a \texttt{GROUP BY} in SQL), or \emph{order-dependent}
(consecutive-transaction gaps, needing an ordered window). To test the substrate and not a deliberately minimal
engine, SQL is granted the median and population-standard-deviation aggregates SQLite omits, defined to match the
DSL; each query is also capped at 5 seconds of wall-clock (correct aggregates run in well under 1 second), so a
pathological model query is scored wrong rather than stalling the evaluation. Because the inner-loop set has few order-dependent
questions, we add eight curated gaps probes (longest, shortest, mean, median, and standard deviation of the
inter-transaction gaps, and so on), all exact in the DSL and all needing a window function in SQL. For each
question the model writes five SQLite queries; we execute each on a per-merchant table and take the majority
value (no gold peeking), then compare to the DSL result. Accuracy is per query, each (question, merchant) instance
scored against gold, the same convention as the BookSQL and EHRSQL baselines; an authored reference SQL reproduces
gold on all order-dependent instances (337/337), so every question is answerable in SQL and each miss is a model
error. Table~\ref{tab:sql} reports the result, including a three-exemplar few-shot column that lifts the
order-dependent category from 0.61 to 0.79 (model-dependent; see BookSQL and EHRSQL for backbones the same exemplars
regress). The order-dependent \emph{failures} are typically both wrong and long; a representative one for ``the sum of all inter-transaction gaps in hours'',
\texttt{sum(gaps\_hours(timestamps))} in the DSL, is:
\begin{lstlisting}
SELECT SUM(gaps_hours) FROM (
  SELECT (julianday(t2.timestamp) - julianday(t1.timestamp)) * 24 AS gaps_hours
  FROM txns t1 JOIN txns t2 ON t1.rowid = (
    SELECT rowid FROM txns WHERE timestamp > t1.timestamp
    ORDER BY timestamp LIMIT 1) ...
\end{lstlisting}
which is not only verbose but incorrect here, since \texttt{julianday} expects a date string while
\texttt{timestamp} holds unix seconds.

\input{tables/sql_baseline}

\section{Reach of the admission gate}
\label{app:reach}
A draft operator must pass two filters in turn: the method-free \emph{substrate} sandbox
(Appendix~\ref{app:sandbox}), then the oracle-free \emph{verification} gate (Appendix~\ref{app:gate}). The gate's
reach is their intersection: what a method-free body can express, and what a conservation relation can vouch for.
We bucket all 384 drafts the model authored across the six induction runs by which filter decided each
(Table~\ref{tab:reach}).

Two things follow. First, throughput is bounded by the substrate, not by verification. The method-free sandbox
turns away 70\% of drafts, almost always because the model reached for a method call on a harder operator, while
verification turns away 17\%. The modest admission yield therefore comes from a deliberate choice to keep the
sandbox auditable; relaxing it would raise yield without touching soundness. Second, verification is precise as
well as sound: every draft it rejects fails a real check, either a metamorphic relation or the differential test,
not a bookkeeping formality. Together with the zero false admits of the mutation study
(Appendix~\ref{app:mutation}), the gate admits no wrong operator and rejects few correct ones.

The capability map is what one would predict. Per-group reductions, counts, and distinct-counts each carry a
parts-to-whole conservation law, and these are the classes that get in. The most-requested capability, a
\emph{mode} (the most-frequent value), is admitted by neither filter: its drafts die at the substrate, since
computing a mode wants a method call, and even written method-free it is an argmax over keys with no conservation
law. Its \emph{count}, the frequency of the mode, does pass, because group counts sum to the number of records.
The reach is thus predictable in advance: conservation-checkable reductions are inside it, non-numeric selections
are not.

\input{tables/gate_reach}

\section{Base-model ablation: is self-extension specific to the 35B?}
\label{app:model-ablation}
The main results run on one base model. To test whether self-extension is a property of that particular model
or of the method, we re-run the entire closed loop, inner loop, induction, and gate, unchanged, with four dense
models from the same family and generation (Qwen3-4B, 8B, 14B, and 32B) on the same three starting sets and the
same fixed 63-merchant validation set, with gate-topk $K=3$ (Appendix~\ref{app:algo}) so a seed continues to a
further induction round whenever the prior round admits an operator. Figure~\ref{fig:seed-model-ladder} and
Table~\ref{tab:model-ablation} report the outcome; the Qwen3.6-35B-A3B used in the main experiments is shown
alongside as a reference point (its numbers are the closed-loop run reported in Table~\ref{tab:closed}). The
dense sweep isolates capability within one family, generation, and architecture, varying only size; the
35B-A3B is a mixture-of-experts of a later generation with far fewer active parameters, so we read it as a
cross-architecture reference point rather than a rung of the same parameter-count axis.

\input{figures/seed_model_ladder}
\input{tables/base_model_ablation}
\input{tables/cross_family}

Two things hold across the ladder, and one does not. \emph{The gate stays sound at every scale.} No false
operator is admitted in any of the seventeen runs (four dense models $\times$ three seeds, plus five continuation
rounds: 4B, 14B, and 32B on amounts/timing, and 32B again on timing and top-buyer). Every admitted operator compiles in the method-free
sandbox and passes both differential testing and all of its declared metamorphic relations. \emph{The inner loop
works at every scale}: even the smallest model here, 4B, expands coverage past the ten seed types on all three
sets (to 39, 17, and 36), and 8B matches or exceeds it on every seed
(57, 36, 42) despite admitting no operators at all. \emph{What does not hold is a monotone relationship between
size and induction yield.} 8B's $128$--$192$ drafts per seed are turned away by the gate's own checks on every
one of the three starting sets. 8B's failure is
not simply a harsher version of the sandbox bottleneck every model hits: of its $384$ drafts across the three
seeds, $80$ ($21\%$) compile under the method-free sandbox, somewhat below the main model's $30\%$ aggregate
substrate pass rate (Appendix~\ref{app:reach}), but among those $80$ compiling drafts, $70$ ($87.5\%$) never
produce a value on the real validation records at all, versus a $6\%$ share for that same failure mode in the
main model's pool (Appendix~\ref{app:reach}); only $9$ fail a declared relation and $1$ a differential check,
and none are admitted. So 8B is not unusually bad at writing method-free code; among the drafts that do compile,
it is unusually bad at writing code that runs at all on real data, a distinct and sharper failure than the
style-compliance bottleneck that dominates elsewhere in this paper. 4B admits four
operators on the amounts/timing seed (gap \texttt{group\_by}, then \texttt{mode} on a second round) but they
unlock no further measured coverage within this run's budget; 14B's five induced operators are used by $17$
further types, and 32B's eight by $50$. Every dense model that
self-extends does so on amounts/timing; the largest, 32B, is the only one to self-extend beyond it, on both timing (three
operators used by $7$ types) and top-buyer (two used by $23$). The 35B-A3B main model, with far fewer active
parameters than the largest dense model here, self-extends on all three seeds ($8$, $9$, and $6$ operators
admitted, summed across both induction rounds; Figure~\ref{fig:seed-model-ladder}). Self-extension is a property of
the method, and the gate's soundness holds regardless of model capability. But induction \emph{yield} does not
simply track parameter count within one family: what determines whether a model self-extends is what it can
express and defend under the gate's checks, separate from raw size.

\paragraph{The same test on the two transfer domains.} We repeat this ablation on BookSQL and EHRSQL
(Appendices~\ref{app:booksql},~\ref{app:ehrsql}): four dense models (4B/8B/14B/32B) against each domain's
own single seed, with the domain-transfer 35B-A3B run as the reference point
(Figure~\ref{fig:model-ladder}). The property we most want to hold on unfamiliar schemas does: the gate
stays sound in both domains, no admitted operator fails differential testing or its own declared relations, and
where a model fails to induce it fails by writing nothing admissible, never by having something wrong slip
through. Induction also fires across the size range rather than only at the top. On BookSQL every dense model
induces successfully. On EHRSQL, only 32B induces (closing $13 \to 139$) while 4B, 8B, and
14B each independently converge on the same top-ranked gap, \texttt{mode}, and are rejected on every attempt,
mostly by the sandbox (a reverted method call, the dominant rejection cause throughout this paper) but also by a
failed metamorphic relation or disagreement between the two implementations, leaving them at 111, 80, and 98
purely from the inner loop on the nine seed operators.

\paragraph{Reading the ladder: coverage height is volume-sensitive.} We read
Figure~\ref{fig:model-ladder} for gate soundness and for whether induction fires. Absolute inner-loop
coverage on these two schemas is a poor stand-in for a capability ranking across models: both
expose many scalar metadata fields (transaction counts, record spans, distinct-party counts, patient ages), and
a weaker model that emits a larger, less canonical spray of candidate expressions over them accumulates more
distinct round-trip-verified value-signatures, so coverage height partly tracks generation volume, which scrambles
the ordering here: for instance 4B reaches 111 answerable types on EHRSQL, above both 8B's 80 and 14B's 98,
whereas on the narrower merchant schema (Figure~\ref{fig:seed-model-ladder}), where there is little to spray, it
does not lead. The effect is bounded and does not reach the growth claims: dropping the handful
of data-independent (constant-valued) expressions moves the main-run totals by under $2\%$ ($15 \to 162$ becomes
$15 \to 159$ on BookSQL, $13 \to 248$ becomes $13 \to 246$ on EHRSQL), and additionally collapsing every trivial
monotone variant (unit rescalings, reciprocal ratios) leaves $15 \to 141$ and $13 \to 239$, so every reported
growth multiple is robust to the counting convention.

\section{Generalization to a second domain: BookSQL}
\label{app:booksql}
The main results run on one domain, merchant transaction QA. To test whether the method generalizes to a
schema built around different entities and fields, not just a different model, we ran the unmodified inner and
outer loop on \textbf{BookSQL}~\citep{kumar2024booksql}, a public business-accounting dataset unrelated to the merchant corpus: entities
are ledgers, one per \texttt{(business, account)} pair, and events are billed transactions in \$ rather than
customer purchases. Unlike the main corpus, BookSQL is real, publicly released data (not a synthetic stand-in);
this is a different concern from the main text's synthetic-corpus motivation, which was specifically about not
exposing proprietary customer data, and does not apply to a dataset that is already public.

\paragraph{What a record is.} One entity per ledger: an account name, a business identifier, the ledger's date
span, its total entry count, and its chronological list of billed entries (date, weekday, amount, entry type,
and counterparty id/name/role). The loop reuses the same restricted namespace and seed operators as the
merchant study; no new harness code was needed, only a schema-adapting parser.

\paragraph{Stats.} 1,753 accounts across 27 businesses. Entry counts per account are heavily skewed (median 1,
mean 462, max 6,213; 945 of 1,753 accounts are single-entry, a structural property of certain account types
rather than noise), so the usable pool for a validation cohort is the 754 accounts with at least 30 entries.

\paragraph{Result.} On a 63-account cohort (matching the main study's validation-set size), seeded from 15
ledger-native question types and the same 9 base operators used for the merchant amounts/timing seed, the
closed loop ran two full induction rounds with no round self-limiting:
\begin{itemize}
\item Inner loop on the 9 seed operators: $15 \to 73$ answerable types over 8 passes.
\item Induction 1 admits 4 operators (\texttt{mode}, \texttt{mode\_grouped}, \texttt{mode\_group},
\texttt{mode\_per\_group}), closing a most-common-value gap.
\item Inner loop on the enlarged 13-operator library: $73 \to 136$ over 8 more passes; $8/63$ of this phase's
newly answerable types call an induced operator (e.g.\ \texttt{mean(mode\_grouped(buyer\_ids, amounts))} for
``the average amount of the most frequent transaction for each buyer'').
\item Induction 2 admits 3 more operators (\texttt{mean\_group}, \texttt{mean\_per\_group},
\texttt{mean\_grouped}), closing a grouped-mean gap.
\item Inner loop on the enlarged 16-operator library: $136 \to 162$ over 8 more passes; $15/26$ of this phase's
newly answerable types call an induced operator.
\end{itemize}
The answerable set grows more than tenfold from the 15 seed types, across the same admission gate, unmodified,
on a domain the method was never designed around. A base-model ablation for this domain (does self-extension
depend on the 35B specifically?) is in Appendix~\ref{app:model-ablation}.

\paragraph{Text-to-SQL comparison.} Scored per query, the same convention as the merchant (Table~\ref{tab:sql}) and
EHRSQL (Table~\ref{tab:ehrsql}) baselines. The pattern matches those baselines: flat aggregation is easy for every
model at scale (\emph{agg} is at or near ceiling from Qwen3-4B up), while the order-dependent category is the hardest
zero-shot for five of the six models (the exception, Qwen3-8B, finds the derived-ratio category marginally harder). The
few-shot column tells the same story as the other domains: three form-only exemplars sharply raise order for the larger
Qwen models (32B and 35B to $\approx$1.00) but the effect is model-dependent, regressing Qwen3-8B and Gemma. The DSL is
exact by construction across every category and every model.

\input{tables/booksql_baseline}

We also run the full closed loop, unmodified, on a third domain, de-identified clinical encounter records
(EHRSQL); see Appendix~\ref{app:ehrsql} for the dataset, the text-to-SQL fairness comparison, and the
self-extension result.

\section{Generalization to a third domain: EHRSQL}
\label{app:ehrsql}
The main results run on merchant transaction QA and the second-domain study runs on business ledgers. To test
the method on a schema built around yet another kind of entity, we run it on \textbf{EHRSQL}~\citep{lee2022ehrsql}, over MIMIC-III~\citep{johnson2016mimic},
where one entity is one patient (\texttt{SUBJECT\_ID}). Unlike the synthetic merchant corpus, MIMIC-III is real,
de-identified clinical data available only to credentialed researchers under PhysioNet's Data Use Agreement. It
is an established external benchmark rather than a synthetic stand-in, so the main text's synthetic-corpus
motivation, which concerns not exposing proprietary customer data, does not apply here; we instead comply with
MIMIC-III's own governance directly, using it under credentialed access.

\paragraph{What a record is.} One entity per patient: that patient's diagnoses, procedures, lab tests, and
prescriptions merged into a single chronological event stream, together with per-visit fields (admission type,
insurance, marital status, ethnicity) and stay durations, so every question is answerable from one entity with
no cross-table join. The loop reuses the same restricted namespace and seed operators as the merchant study; no
new harness code was needed, only a schema-adapting parser.

\paragraph{Stats.} 300 patients. Two evaluation tracks are built and round-trip verified (below): 7,660 verified
compositional instances and 4,500 verified retrieval instances, both scored in full here.

\paragraph{The fairness setup.} EHRSQL's published gold SQL is written against the normalized 17-table
MIMIC-III schema, so most questions require JOINs the DSL never performs. We remove that asymmetry exactly as
for the merchant study: text2sql answers over a \emph{flat per-patient table that mirrors the DSL context
one-to-one}, so it writes single-table queries and the comparison isolates NL$\to$answer reasoning, not schema
navigation. Gold is trustworthy by round-trip verification: for each instance we compute the answer two
independent ways and keep it only if they agree (for retrieval questions, the original gold SQL on the real DB
vs.\ the value derived from the DSL context; for compositional questions, the DSL evaluator vs.\ an independent
flat-SQL query). As in Table~\ref{tab:sql}, SQL is granted the median and population-standard-deviation
aggregates SQLite lacks. Length of stay is recomputed with \texttt{julianday} (the published day-of-year gold
is off by fractional days and negative across year boundaries), a documented deviation from the EHRSQL
leaderboard confined to the two duration templates.

\paragraph{Two tracks.} The main track is \emph{Compositional}: self-generated questions that stress the DSL's
operators (standard deviation, median, coefficient of variation, consecutive-event gaps, category proportions),
with gold agreed by the two independent references above and no external answer key. A supplementary
\emph{Retrieval} track of real EHRSQL attribute look-ups acts as a fairness control: the DSL is exact on both
tracks, and five of the six text-to-SQL baselines score higher on Retrieval than on Compositional (Table~\ref{tab:ehrsql}),
so the compositional gap is better explained by reasoning difficulty than by a baseline that is weak across the board.

\paragraph{Scoring.} We score per query, i.e., each (question, patient) instance against gold under the same
denotation tolerance as the merchant and BookSQL studies. EHRSQL questions are patient-specific over 300 patients,
so an all-or-nothing-across-300 unit would collapse almost every template on a single hard patient; per-query is
also how EHRSQL is scored in the benchmark literature. An authored flat-SQL reference reproduces gold across the
compositional pool by the round-trip verification above, so every text2sql miss is a genuine model failure, not a
mislabeled question.

\paragraph{Text-to-SQL comparison.} The text2sql baseline across six open-weight backbones (best-of-5,
temperature 0.4) is Table~\ref{tab:ehrsql}. Within the compositional track the failures concentrate in the
order-dependent category (consecutive-event gaps needing an ordered window function), the same category that is
hardest zero-shot in the merchant and BookSQL studies; the median/stddev grant makes the plain aggregates perfect
across all models. A three-exemplar few-shot column lifts every Qwen backbone but is model-dependent: Gemma-3-27B
drops, because the window exemplar induces an aggregate-over-window SQLite rejects where zero-shot it used legal
self-joins. The DSL is exact by construction on both tracks.

\input{tables/ehrsql_baseline}

\paragraph{Result.} On a 63-patient cohort (matching the main study's validation-set size), seeded from 13
clinically-native question types and the same 9 base operators used for the merchant amounts/timing seed, the
closed loop runs two full induction rounds with no round self-limiting. The inner loop on the 9 seed operators
grows the answerable set $13 \to 138$ over 8 passes. Induction 1 closes a most-common-value gap, admitting 17
distinct operators (\texttt{mode}, \texttt{per\_group\_sum}, \texttt{grouped\_sum}, \texttt{group\_reduce},
\texttt{group\_agg}, among others); many independently-sampled drafts converge on the same per-group-reduction
capability under different names (e.g.\ \texttt{sum\_by}, \texttt{agg\_by}, \texttt{by\_group},
\texttt{keyed\_reduce}), and the harness registers every distinct name that independently passes the gate, not
only the first. The inner loop on the enlarged 26-operator library grows the set $138 \to 222$ over 8 more
passes; $48/84$ of this phase's newly answerable types call an induced operator, e.g.\
\texttt{max(grouped\_agg(dept\_codes, amounts, 'sum'))} for ``the maximum total cost among all department
types (diagnoses, procedures, labs, and prescriptions) billed for this patient.'' Induction 2 closes a
filtered-sum gap, admitting 9 more operators (\texttt{cond\_agg}, \texttt{conditional\_sum},
\texttt{conditional\_agg}, \texttt{filter\_sum}, \texttt{threshold\_sum}, among others). The inner loop on the
enlarged 35-operator library grows the set $222 \to 248$ over 8 more passes; $9/26$ of this phase's newly
answerable types call an induced operator, e.g.\ \texttt{mean(conditional\_agg(amounts, dept\_codes, "LAB",
"=="))} for ``the average cost of this patient's lab tests.'' The answerable set grows nearly twentyfold from
the 13 seed types, across the same admission gate, unmodified, on a domain built around individual patients
rather than merchants or business ledgers. A base-model ablation for this domain is in
Appendix~\ref{app:model-ablation}.

\section{Convergence: independent seeds reach a shared per-group core}
\label{app:convergence}
The three starting sets begin with disjoint themes and different minimal operator sets
(Appendix~\ref{app:seeds}), so a natural question is whether they merely grow in separate directions or converge.
We measure convergence directly. Every answerable type is recorded with its value signature, the vector of its
expression's values over the fixed 63-merchant validation set (Appendix~\ref{app:repro}); because all three runs
use the same 63 merchants, two types from different runs compute the same quantity exactly when their signatures
match. We can therefore intersect the runs' answerable sets by quantity, not by surface form. This value
signature is also how we identify \emph{semantic duplicates across independently-named operators}: two operators
are the same capability when they induce the same signature on the shared merchants, so the ``distinct operators
admitted'' counts in Section~\ref{sec:results} collapse name variants to one. (Analysis script:
\texttt{convergence.py} over the committed run artifacts; numbers in \texttt{convergence\_results.json}.)

Figure~\ref{fig:convergence} plots the resulting 3-way intersection across induction stages. Table~\ref{tab:convergence-ops}
shows the sharpest form of that convergence at the operator level: $13$ of the shared-core quantities are each
computed by a differently-named per-group operator in every run, and the table lists four representative rows.

\input{figures/convergence}
\input{tables/convergence_operators}

\section{Operator families induced}
\label{app:families}
The 16 distinct operators admitted across the main runs (the 14 in the round-1 pool used by the mutation study,
plus \texttt{filter\_and\_reduce} and \texttt{filter\_col}, admitted only in the amounts/timing seed's round-2
chaining) fall into four capability families, not per-group
aggregation alone (classified by declared name and spec; \texttt{operator\_families.py} over the committed
induce artifacts). \emph{Per-group aggregation} (6): \texttt{cond\_sum, group\_by\_and\_reduce, group\_reduce,
group\_reduce\_by\_key, group\_sum, per\_group\_sum}. \emph{Filter, select, rank, and bin} (6), which are not
per-group aggregations: \texttt{conditional\_agg, conditional\_sum, filter\_and\_reduce, filter\_col,
mask\_amounts, per\_filter\_sum}. \emph{Distinct and cardinality} (3): \texttt{per\_group\_count,
per\_group\_reducer, per\_group\_size}. The remaining operator, \texttt{per\_hour\_count}, is a temporal-count
reducer outside the other three families. The gate's reach is thus not tied to a single operator family.

\section{Independence of the reference: separate call and cross-family}
\label{app:independence}
The differential reference is written independently of the implementation. We stress this independence at three
levels. \textbf{Co-generated} (baseline): implementation and reference emitted in one response; the three
starting sets admit $6$, $5$, and $2$ operators. \textbf{Separate call, same model}: the reference is generated
in a distinct call conditioned only on the operator's name, signature, and one-line spec, never shown the
implementation. Re-running the full natural induction this way (\texttt{induce.py --separate-reference};
\texttt{sepref\_induce\_\{amt,buyer,timing\}\_results.json}) still self-extends on every set, admitting $7$, $7$,
and $4$ operators (discovering mode, where, and mode as the respective gaps). \textbf{Different model family}: the
reference is written by Gemma-3-27B while the implementations come from the Qwen base model. On the mutation pool
(\texttt{mutation\_soundness.py --separate-reference --llm-path .../gemma-3-27b-it};
\texttt{sepref\_mutation\_crossfam\_results.json}) the cross-family reference admits $0/138$ buggy mutants under
an independent reference (and $10/138$ under the correlated worst-case, as with the same-family reference).
Independence therefore does not
depend on co-generation or even on sharing a model.

\section{Repeated-run variance}
\label{app:variance}
Because sampling is the only source of randomness, we repeat the full outer loop five times per starting set
(\texttt{repeat\_variance.py}; \texttt{repeat\_variance\_results.json}). Admitted-operator counts are
amounts/timing $\{22,19,5,5,19\}$ (mean $14\pm7.43$), top-buyer $\{2,3,3,3,1\}$ ($2.4\pm0.8$), and timing
$\{5,6,4,2,6\}$ ($4.6\pm1.5$). At least one operator is admitted in every run across all three starting sets
($5/5$ each). The stable statistic is the direction (a positive count, versus zero for the no-operator
control), not the exact number.

\section{False-rejects among naturally-proposed operators}
\label{app:false-reject}
The $1/14$ false-reject figure is measured on the 14 admitted operators; a fairer question is how often the gate
rejects a \emph{correct} operator among the ones the model actually proposes. We measure this oracle-free over
the committed induction drafts (\texttt{false\_reject\_analysis.py}; \texttt{false\_reject\_results.json}). Of
$192$ drafts, $17$ are admitted and $175$ rejected. The rejections break down as $141$ ($80.6\%$) sandbox
violations (the reference or implementation used a forbidden method or failed to compile), $9$ ($5.1\%$)
differential disagreements (the two implementations differ, so at least one is wrong), $22$ ($12.6\%$)
relation-only (the two implementations agree but a declared metamorphic relation failed), and $3$ other. Within
the relation-only bucket, recomputing each draft's value signature on the synthetic merchants shows $10$ whose
signature equals an already-admitted operator: correct operators rejected only because the model mis-stated a
conservation law. These are a throughput cost, not a lost capability, since the same quantity is admitted through
another draft.

\section{Admission-rule baselines}
\label{app:admission-baselines}
We compare the admission rule against the teacher-free signals self-evolving agents and library-learning systems
rely on, measured on the same $138$ buggy mutants as the mutation study (\texttt{admission\_baselines.py};
\texttt{admission\_baselines\_results.json}), plus the LLM-judge rule (Appendix~\ref{app:judge}).
\begin{center}
\small
\begin{tabular}{lc}
\toprule
Admission rule & Buggy admitted \\
\midrule
Admit-if-it-runs (reward/compression floor) & $129/138$ \\
Self-consistency (operator agrees with itself) & $129/138$ \\
Reuse / compression (call shape recurs) & $101/138$ \\
LLM-judge & $0/138$ admitted, but $2/14$ correct \emph{rejected} \\
Independent agreement + conservation (ours) & $\mathbf{0/138}$, and $1/14$ correct rejected \\
\bottomrule
\end{tabular}
\end{center}
Each teacher-free signal is correctness-blind for a self-authored operator: a wrong operator runs, is
self-consistent, and shares the correct operator's call shape. Requiring agreement with an \emph{independent}
implementation plus conservation is what separates correct from wrong.

\section{Human validation: protocol and agreement}
\label{app:human}
As an external counterpart to the automated held-out check (\S\ref{sec:inner}), three annotators (members of the
project team) independently rated a seeded, stratified sample of the agent's self-generated outputs, blind to the
gate's admission decisions and to one another: $150$ question and DSL-expression pairs (stratified across the
three domains and the seed- and induced-operator segments) and all $53$ distinct induced operators. For each
question they judged whether the DSL expression computes exactly what the question asks (faithfulness) and whether
the question is well-formed; for each operator they first wrote a description of the implementation blind to its
name, then judged whether the two independently generated implementations are equivalent, whether the revealed
name and specification match the code, and whether the operator is a sensible reusable primitive.

By majority vote, $89\%$ of expressions were judged faithful and $97\%$ of questions well-formed; faithfulness was
higher on seed-operator questions ($93\%$) than on the harder induced-operator questions ($84\%$). The three
annotators agreed on $91\%$ of pairwise faithfulness judgments. Because the faithful label dominates (about $88\%$
prevalence), chance-corrected agreement is better summarized by Gwet's AC$_1$, which is robust to high prevalence
\citep{gwet2008}, than by Fleiss' or Cohen's $\kappa$, which that prevalence deflates: AC$_1 = 0.88$ for
faithfulness and $0.94$ for well-formedness. On the $53$ induced operators, the annotators unanimously judged
every pair of independent implementations equivalent (the property the gate certifies automatically,
\S\ref{sec:soundness}), and both annotators judged the revealed name and specification accurate, and the operator
a sensible primitive, for $81\%$ of operators on each axis.

%% file: algo/self_extend.tex
\begin{algorithm}[t]
\caption{Self-extending DSL agent. Both loops run on a frozen base model $\pi$ with no ground-truth oracle and no
LLM judge. $\textsc{sig}(e,M)$ is expression $e$'s value vector over the validation records $M$; equal signatures
mean equal quantities. $\textsc{Metamorphic}$ checks conservation and structural relations stated over the trusted
primitives (Appendix~\ref{app:gate}).}
\label{alg:selfextend}
\small
\begin{algorithmic}[1]
\Procedure{InnerLoop}{starting question types $T$, operators $O$, records $M$}
  \State $K \gets \{\, \textsc{sig}(t, M) : t \in T \,\}$ \Comment{quantities already answerable}
  \Repeat
    \For{each $(q, e)$ the model proposes as a (question, expression) pair using only $O$}
      \State $s \gets \textsc{Evaluate}(e, M)$ \Comment{skip if $e$ uses an operator $\notin O$ or is not a single number}
      \State $E' \gets \textsc{AnswerFromQuestion}(q)$ \Comment{expressions written from the question text alone}
      \If{some $e' \in E'$ has $\textsc{sig}(e', M) = s$ \textbf{and} $s \notin K$}
        \State add $s$ to $K$; \; record the question type $(q, e)$ \Comment{agreement + not seen before}
      \EndIf
    \EndFor
  \Until{a full pass adds nothing to $K$}
  \State \Return the recorded question types
\EndProcedure
\Statex
\Procedure{OuterLoop}{operators $O$, records $M$, a request outside the closure of $O$}
  \State $(f, f') \gets \textsc{Induce}(\text{missing capability})$ \Comment{model writes TWO independent implementations}
  \State $ok \gets \textsc{Differential}(f, f', M) \ \wedge \ \textsc{Metamorphic}(f, M)$ \Comment{agree on every input; obey trusted relations}
  \If{$ok$}
    \State \Return $\textsc{InnerLoop}(\text{starting types}, \ O \cup \{f\}, \ M)$ \Comment{admit; downstream round-trip re-checks each type}
  \Else
    \State \Return reject $f$ \Comment{admitting it would break exact verifiability}
  \EndIf
\EndProcedure
\end{algorithmic}
\end{algorithm}

%% file: figures/seed_trajectory.tex
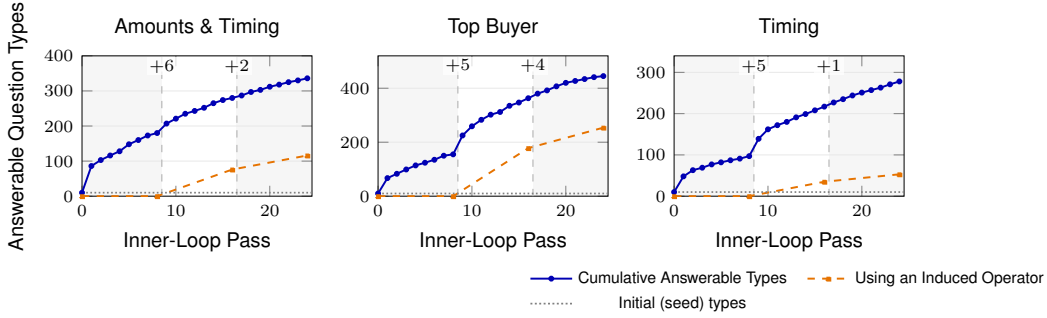
\begin{figure}[t]
\centering
\resizebox{\linewidth}{!}{%
\begin{tikzpicture}
\begin{groupplot}[
  group style={group size=3 by 1, horizontal sep=0.95cm},
  width=0.35\linewidth, height=3.6cm,
  xlabel={Inner-Loop Pass}, xmin=0, ymin=0,
  ymajorgrids=true, grid style={gray!18, line width=0.3pt},
  tick label style={font=\sffamily\scriptsize}, label style={font=\sffamily\footnotesize},
  title style={font=\sffamily\footnotesize, yshift=-2pt}, every axis plot/.append style={line width=0.8pt},
  legend style={font=\sffamily\scriptsize, draw=none, at={(0.5,-0.46)}, anchor=north, legend columns=2, /tikz/every even column/.append style={column sep=6pt}}]

\nextgroupplot[title={Amounts \& Timing}, ylabel={Answerable Question Types}, ymax=400, xmax=24.5]
\begin{scope}[on background layer]
  \fill[gray!7] (axis cs:0,0) rectangle (axis cs:8.5,400);
  \fill[gray!7] (axis cs:16.5,0) rectangle (axis cs:24.5,400);
\end{scope}
\draw[gray!55,dashed,line width=0.5pt] (axis cs:8.5,0)--(axis cs:8.5,400);
\draw[gray!55,dashed,line width=0.5pt] (axis cs:16.5,0)--(axis cs:16.5,400);
\node[font=\sffamily\scriptsize,anchor=north,inner sep=1pt,fill=white,fill opacity=0.75,text opacity=1] at (axis cs:8.5,398) {$+6$};
\node[font=\sffamily\scriptsize,anchor=north,inner sep=1pt,fill=white,fill opacity=0.75,text opacity=1] at (axis cs:16.5,398) {$+2$};
\addplot[blue!70!black,mark=*,mark size=0.8pt] coordinates {(0,10) (1,86) (2,103) (3,116) (4,128) (5,148) (6,160) (7,173) (8,180) (9,207) (10,221) (11,235) (12,243) (13,252) (14,265) (15,274) (16,280) (17,287) (18,297) (19,303) (20,312) (21,318) (22,325) (23,330) (24,336)};
\addplot[orange!90!black,dashed,mark=square*,mark size=0.8pt] coordinates {(0,0) (8,0) (16,76) (24,116)};
\draw[densely dotted, black!50, line width=0.7pt] (axis cs:0,10) -- (axis cs:24.5,10);

\nextgroupplot[title={Top Buyer}, ymax=520, xmax=24.5]
\begin{scope}[on background layer]
  \fill[gray!7] (axis cs:0,0) rectangle (axis cs:8.5,520);
  \fill[gray!7] (axis cs:16.5,0) rectangle (axis cs:24.5,520);
\end{scope}
\draw[gray!55,dashed,line width=0.5pt] (axis cs:8.5,0)--(axis cs:8.5,520);
\draw[gray!55,dashed,line width=0.5pt] (axis cs:16.5,0)--(axis cs:16.5,520);
\node[font=\sffamily\scriptsize,anchor=north,inner sep=1pt,fill=white,fill opacity=0.75,text opacity=1] at (axis cs:8.5,518) {$+5$};
\node[font=\sffamily\scriptsize,anchor=north,inner sep=1pt,fill=white,fill opacity=0.75,text opacity=1] at (axis cs:16.5,518) {$+4$};
\addplot[blue!70!black,mark=*,mark size=0.8pt] coordinates {(0,10) (1,67) (2,83) (3,99) (4,114) (5,124) (6,135) (7,150) (8,155) (9,225) (10,259) (11,283) (12,302) (13,312) (14,335) (15,347) (16,363) (17,380) (18,392) (19,407) (20,420) (21,427) (22,434) (23,441) (24,445)};
\addplot[orange!90!black,dashed,mark=square*,mark size=0.8pt] coordinates {(0,0) (8,0) (16,178) (24,254)};
\draw[densely dotted, black!50, line width=0.7pt] (axis cs:0,10) -- (axis cs:24.5,10);

\nextgroupplot[title={Timing}, ymax=340, xmax=24.5]
\begin{scope}[on background layer]
  \fill[gray!7] (axis cs:0,0) rectangle (axis cs:8.5,340);
  \fill[gray!7] (axis cs:16.5,0) rectangle (axis cs:24.5,340);
\end{scope}
\draw[gray!55,dashed,line width=0.5pt] (axis cs:8.5,0)--(axis cs:8.5,340);
\draw[gray!55,dashed,line width=0.5pt] (axis cs:16.5,0)--(axis cs:16.5,340);
\node[font=\sffamily\scriptsize,anchor=north,inner sep=1pt,fill=white,fill opacity=0.75,text opacity=1] at (axis cs:8.5,338) {$+5$};
\node[font=\sffamily\scriptsize,anchor=north,inner sep=1pt,fill=white,fill opacity=0.75,text opacity=1] at (axis cs:16.5,338) {$+1$};
\addplot[blue!70!black,mark=*,mark size=0.8pt] coordinates {(0,10) (1,48) (2,63) (3,69) (4,77) (5,82) (6,87) (7,91) (8,97) (9,139) (10,162) (11,172) (12,180) (13,191) (14,199) (15,208) (16,217) (17,227) (18,235) (19,244) (20,251) (21,257) (22,263) (23,271) (24,278)};
\addplot[orange!90!black,dashed,mark=square*,mark size=0.8pt] coordinates {(0,0) (8,0) (16,35) (24,53)};
\draw[densely dotted, black!50, line width=0.7pt] (axis cs:0,10) -- (axis cs:24.5,10);
\addlegendimage{densely dotted, black!50, line width=0.7pt}
\legend{Cumulative Answerable Types, Using an Induced Operator, Initial (seed) types}
\end{groupplot}
\end{tikzpicture}%
}
\caption{One continuous closed-loop trajectory per starting set, on a fixed validation set (63 merchants). Shaded
and unshaded bands mark successive inner-loop phases; a dashed vertical line at each phase boundary is an outer-loop
induction, labeled with the number of operators it admits ($+n$). The inner loop
climbs on its current operators (solid), an induction enlarges the library, and it climbs again. The capability
each induction closes, in order, is: a most-common-value gap closed by a family of per-group aggregation and
counting operators, then a related weighted-mode gap closed by more per-group counting operators (Amounts \&
Timing); a count gap closed by conditional-filter and per-group count/sum operators, then an equality gap closed
by equality-filter and grouped-sum operators (Top Buyer); and a most-common-value gap closed by per-group counting
operators, then an argmax gap closed by a single grouped-sum reducer (Timing). The orange dashed
line counts only the answerable types whose expression \emph{calls an operator admitted this round}: it is flat at
zero through the first inner phase (no operator to use) and rises only after an induction, so the post-induction
gains are genuinely induction-driven, not the inner loop continuing on the base operators. Of the newly answerable
types in the first post-induction phase, $76/100$ (Amounts \& Timing), $178/208$ (Top Buyer), and $35/120$ (Timing)
call an induced operator; the cumulative counts after the second post-induction phase are $116$, $254$, and $53$
respectively. All three starting sets admit new operators at both inductions and keep climbing through a third
inner phase; none self-limits within the two induction rounds tested here. Inner phases are capped at eight
passes and are still rising at the cap, so these are lower bounds on each phase's reachable set.}
\label{fig:seed-trajectory}
\end{figure}

%% file: tables/saturation.tex
\begin{table}[t]
\caption{Inner-loop coverage growth on the amounts/timing starting set with a fixed operator set (diminishing
returns; each round counts only types not found earlier). Columns: distinct (question, expression) pairs
\emph{proposed}, those \emph{executable} to one number per record, those also \emph{verified} by the
independent-expression check, the \emph{new} types, and the cumulative \emph{total}. The ten seed types count in
\emph{total} but not \emph{new} (\S\ref{sec:inner}).}
\label{tab:saturation}
\centering
\small\renewcommand{\arraystretch}{0.95}
\begin{tabular}{ccccccc}
\toprule
Round & Known before & Proposed & Executable & Verified & New & Total answerable \\
\midrule
1 & 0  & 82 & 40 & 35 & 26 & 36 \\
2 & 26 & 75 & 48 & 40 & 18 & 54 \\
3 & 44 & 68 & 43 & 39 & 9  & 63 \\
\bottomrule
\end{tabular}
\end{table}

%% file: tables/closed_loop.tex
\begin{table}[t]
\caption{Automated outer-loop induction on three independent starting sets, each from its \emph{own} minimal
operator set (63 validation merchants). \emph{Using induced} counts new types whose expression calls an operator
admitted this round; the control runs amounts/timing with induction skipped (Appendix~\ref{app:seeds}).}
\label{tab:closed}
\centering
\resizebox{\columnwidth}{!}{%
\begin{tabular}{lcccccccc}
\toprule
Starting set & Init.\ ops & Gap & Admitted & Proposed & Executable & Verified & New & Using induced \\
\midrule
Amounts/timing & 9  & mode  & 6 & 387 & 108 & 79 & 65 & \textbf{25} \\
Top-buyer      & 8  & where & 5 & 590 & 140 & 81 & 62 & \textbf{6}  \\
Timing         & 15 & mode  & 2 & 332 & 133 & 95 & 66 & \textbf{2}  \\
\midrule
Amounts/timing (round 2) & 15 & mode & 2 & 339 & 119 & 88 & 67 & \textbf{3} \\
\midrule
Control (amounts/timing, no induction) & 9 & n/a & 0 & 277 & 103 & 92 & 69 & \textbf{0} \\
\bottomrule
\end{tabular}%
}
\end{table}

%% file: tables/compress.tex
\begin{table}[t]
\caption{Verified compression: the most-reused sub-expression patterns among the 219 verified expressions are
folded into named operators, each admitted only if \emph{equivalent} to the pattern it replaces on every record
(63 merchants). \emph{Sites} is occurrences across \emph{questions} distinct questions; \emph{checks} is the
equivalence evaluations (sites $\times$ merchants) that all agreed.}
\label{tab:compress}
\centering
\begin{tabular}{lcccc}
\toprule
Compressed operator & Sites & Questions & Chars saved & Equiv.\ checks \\
\midrule
\texttt{max(per\_group\_sum(a0, a1))} & 14 & 10 & 84 & 882 \\
\texttt{ratio(sum(a0), nunique(a1))}  &  6 &  6 & 42 & 378 \\
\texttt{max(gaps\_days(a0))}          &  4 &  4 &  8 & 252 \\
\texttt{min(gaps\_days(a0))}          &  4 &  4 &  8 & 252 \\
\texttt{mean(gaps\_days(a0))}         &  4 &  4 & 12 & 252 \\
\bottomrule
\end{tabular}
\end{table}

%% file: tables/mutation_soundness.tex
\begin{table}[t]
\caption{Mutation study of the admission gate. For each admitted operator we generate every single-point mutant
of its implementation, discard the ones that still match the original on every input (equivalent mutants), and
run the rest through the gate. \emph{diff.\ kills} is the uncorrelated model (only the implementation is mutated,
the reference stays correct), where differential testing alone should catch the bug. \emph{metamorphic} and
\emph{structural} kills are the correlated model (the same bug is written into both implementations, so
differential cannot catch it), where only the conservation checks or a no-output rejection can catch it.
\emph{false admits} is $0/138$ under the uncorrelated (independent-reference) model the loop uses, and $10/138$
under the correlated worst-case, where the identical bug is written into both implementations so differential is
blind and only metamorphic conservation remains. Validation on 63 synthetic merchants.}
\label{tab:mutation}
\centering
\setlength{\tabcolsep}{5pt}
\small
\begin{tabular}{lccccc}
\toprule
Operator & Buggy mutants & Diff.\ kills & Metamorphic kills & Structural kills & False admits \\
\midrule
\texttt{per\_group\_reducer}   &  7 &  7 &  6 & 1 & 0 \\
\texttt{group\_reduce\_by\_key} & 11 & 11 & 10 & 1 & 0 \\
\texttt{per\_group\_count}     &  5 &  5 &  5 & 0 & 0 \\
\texttt{per\_group\_size}      &  9 &  9 &  7 & 0 & 2 \\
\texttt{per\_group\_sum}       & 13 & 13 & 11 & 2 & 0 \\
\texttt{group\_reduce}         & 13 & 13 & 11 & 1 & 1 \\
\texttt{per\_filter\_sum}      & 12 & 12 & 12 & 0 & 0 \\
\texttt{conditional\_agg}      & 12 & 12 & 11 & 0 & 1 \\
\texttt{per\_hour\_count}      &  4 &  4 &  4 & 0 & 0 \\
\texttt{cond\_sum}             &  9 &  9 &  9 & 0 & 0 \\
\texttt{mask\_amounts}         &  9 &  9 &  4 & 1 & 4 \\
\texttt{group\_by\_and\_reduce} & 12 & 12 & 11 & 1 & 0 \\
\texttt{conditional\_sum}      &  8 &  8 &  6 & 1 & 1 \\
\texttt{group\_sum}            & 14 & 14 & 12 & 1 & 1 \\
\midrule
Total                    & 138 & 138 & 119 & 9 & 10 \\
\bottomrule
\end{tabular}
\end{table}

%% file: tables/adversarial.tex
\begin{table}[t]
\caption{Adversarial, conservation-preserving wrong operators. Each construction is built to satisfy every
relation \texttt{per\_group\_sum} or \texttt{group\_size} declares (relabeling), or is a control that should not
(merge). Validated against the same gate as the closed loop, under the two mutation-study threat models. The
lone false admit, \texttt{per\_group\_sum} + merge under the correlated model, is bold.}
\label{tab:adversarial}
\centering
\begin{tabular}{llcc}
\toprule
Operator & Construction & Uncorrelated & Correlated \\
\midrule
\texttt{per\_group\_sum} & cyclic (relabel) & admitted & admitted \\
\texttt{per\_group\_sum} & swap (relabel)   & admitted & admitted \\
\texttt{per\_group\_sum} & merge (control)  & rejected & \textbf{admitted} \\
\texttt{group\_size}     & cyclic (relabel) & admitted & admitted \\
\texttt{group\_size}     & swap (relabel)   & admitted & admitted \\
\texttt{group\_size}     & merge (control)  & rejected & rejected \\
\bottomrule
\end{tabular}
\end{table}

%% file: tables/judge_baseline.tex
\begin{table}[t]
\caption{LLM-judge admission baseline versus the oracle-free gate, on the same pool (138 mutants, 6 adversarial
constructions, 14 correct operators; 63 merchants). The judge is the same frozen model, majority vote over 5
reasoning samples per candidate; columns as in Table~\ref{tab:ablation}.}
\label{tab:judge}
\centering
\begin{tabular}{lccc}
\toprule
Admission rule & Buggy & Adv. & False rej. \\
\midrule
Oracle-free gate (uncorrelated)    & 0/138  & 4/6 & 1/14 \\
Oracle-free gate (correlated)      & 10/138 & 5/6 & 1/14 \\
LLM judge (same model, maj.\ of 5) & 0/138  & 0/6 & 2/14 \\
\bottomrule
\end{tabular}
\end{table}

%% file: tables/sql_baseline.tex
\begin{table}[t]
\caption{Text-to-SQL baseline on the verified (question, expression) pairs (43 merchants), Qwen3.6-35B-A3B.
Each question's DSL expression is the exact gold; the model emits a SQLite query (best-of-5, majority vote
over the executed results, temperature 0.4) run on a per-merchant transactions table and compared. To test
the substrate rather than a deliberately minimal engine, SQL is granted the two analytics aggregates SQLite
omits, median and population standard deviation (as Postgres/BigQuery/DuckDB provide), matching the DSL's
definitions. Accuracy is per query (each (question, merchant) instance scored against gold), the same
convention as BookSQL (Table~\ref{tab:booksql-sql}) and EHRSQL (Table~\ref{tab:ehrsql}). An authored reference
SQL reproduces gold on all order-dependent instances (337/337), so every question is answerable in SQL.
Numbers are the mean of 3 seeds (per-cell std $\le 0.01$). Lengths are mean characters of the query; the DSL
is exact by construction. The \emph{3-shot} column prepends three hand-authored (question, SQL) exemplars to
the prompt (one GROUP BY, one ordered window, one granted aggregate), disjoint from the eval and illustrating
form only; it lifts the order-dependent category (0.61$\to$0.79) at a small cost to per-group, confirming the
order gap is largely elicitation rather than a substrate limit.}
\label{tab:sql}
\centering
\small\renewcommand{\arraystretch}{0.95}\setlength{\tabcolsep}{5pt}
\begin{tabular}{lcccccc}
\toprule
category & questions & SQL (0-shot) & SQL (3-shot) & DSL acc. & mean SQL len. & mean DSL len. \\
\midrule
flat            & 129 & 0.87 & 0.88 & 1.00 &  59 & 37 \\
per-group       &  25 & 0.75 & 0.71 & 1.00 & 118 & 45 \\
order-dependent &  29 & 0.61 & 0.79 & 1.00 & 258 & 38 \\
\midrule
overall         & 183 & 0.81 & 0.84 & 1.00 & n/a & n/a \\
\bottomrule
\end{tabular}
\end{table}

%% file: tables/gate_reach.tex
\begin{table}[t]
\caption{Reach of the admission gate over 384 operator drafts from six induction runs. A draft must pass two
filters in turn: the method-free \emph{substrate} sandbox, then \emph{verification} (differential and
metamorphic). Most drafts are turned away by the substrate, not by verification, because the model reaches for a
method call on a harder operator; verification, in turn, rejects almost only wrong operators. Sub-counts are
indented.}
\label{tab:reach}
\centering
\setlength{\tabcolsep}{6pt}
\begin{tabular}{lcc}
\toprule
Outcome & Drafts & Share \\
\midrule
Admitted                                & 41  & 11\% \\
Rejected by substrate (method-free)     & 270 & 70\% \\
\quad Method call or attribute access    & 196 &      \\
\quad Other banned syntax                & 74  &      \\
Rejected by verification                & 64  & 17\% \\
\quad A metamorphic relation fails       & 44  &      \\
\quad The two implementations disagree   & 20  &      \\
Rejected: no output (structural)        & 7   & 2\%  \\
Rejected: unparseable / format          & 2   & 1\%  \\
\bottomrule
\end{tabular}
\end{table}

%% file: figures/seed_model_ladder.tex
\begin{figure}[t]
\centering
\resizebox{\linewidth}{!}{%
\begin{tikzpicture}
\begin{groupplot}[
  group style={group size=3 by 1, horizontal sep=1.0cm},
  width=0.36\linewidth, height=3.6cm,
  ybar stacked, /pgf/bar width=10pt,
  symbolic x coords={4B,8B,14B,32B,35B}, xtick=data, enlarge x limits=0.14,
  ymin=0, ymajorgrids=true, grid style={gray!18, line width=0.3pt},
  tick label style={font=\sffamily\scriptsize}, label style={font=\sffamily\footnotesize},
  title style={font=\sffamily\footnotesize, yshift=-2pt},
  legend style={font=\sffamily\scriptsize, draw=none, at={(0.5,-0.24)}, anchor=north, legend columns=2,
    /tikz/every even column/.append style={column sep=6pt}}]

\nextgroupplot[title={Amounts \& Timing}, ylabel={Answerable Question Types}, ymax=400]
\addplot[fill=blue!30, draw=blue!55!black] coordinates {(4B,10) (8B,10) (14B,10) (32B,10) (35B,10)};
\addplot[fill=blue!70!black, draw=blue!55!black] coordinates {(4B,29) (8B,47) (14B,73) (32B,225) (35B,210)};
\addplot[pattern=north east lines, pattern color=orange!90!black, draw=orange!80!black] coordinates {(4B,0) (8B,0) (14B,17) (32B,50) (35B,116)};
\node[font=\sffamily\scriptsize,anchor=south,text=orange!75!black] at (axis cs:14B,100) {+5};
\node[font=\sffamily\scriptsize,anchor=south,text=orange!75!black] at (axis cs:32B,285) {+8};
\node[font=\sffamily\scriptsize,anchor=south,text=orange!75!black] at (axis cs:35B,336) {+8};

\nextgroupplot[title={Top Buyer}, ymax=520]
\addplot[fill=blue!30, draw=blue!55!black] coordinates {(4B,10) (8B,10) (14B,10) (32B,10) (35B,10)};
\addplot[fill=blue!70!black, draw=blue!55!black] coordinates {(4B,7) (8B,26) (14B,51) (32B,241) (35B,181)};
\addplot[pattern=north east lines, pattern color=orange!90!black, draw=orange!80!black] coordinates {(4B,0) (8B,0) (14B,0) (32B,23) (35B,254)};
\node[font=\sffamily\scriptsize,anchor=south,text=orange!75!black] at (axis cs:32B,274) {+2};
\node[font=\sffamily\scriptsize,anchor=south,text=orange!75!black] at (axis cs:35B,445) {+9};

\nextgroupplot[title={Timing}, ymax=340]
\addplot[fill=blue!30, draw=blue!55!black] coordinates {(4B,10) (8B,10) (14B,10) (32B,10) (35B,10)};
\addplot[fill=blue!70!black, draw=blue!55!black] coordinates {(4B,26) (8B,32) (14B,66) (32B,255) (35B,215)};
\addplot[pattern=north east lines, pattern color=orange!90!black, draw=orange!80!black] coordinates {(4B,0) (8B,0) (14B,0) (32B,7) (35B,53)};
\node[font=\sffamily\scriptsize,anchor=south,text=orange!75!black] at (axis cs:32B,272) {+3};
\node[font=\sffamily\scriptsize,anchor=south,text=orange!75!black] at (axis cs:35B,278) {+6};
\legend{Initial (seed) types, Using a Seed Operator, Using an Induced Operator}
\end{groupplot}
\end{tikzpicture}%
}
\caption{Base-model ablation on the merchant domain across all three starting sets (Amounts \& Timing, Top Buyer,
Timing), the seed-sensitivity companion to the cross-domain ladder of Figure~\ref{fig:model-ladder}. Five base
models per seed, same stacked convention: the light band at the base is the initial seed types (constant across
base models), the solid above it is the additional subset expressible with the seed operators alone, hatched is
the subset whose expression calls an induced operator, and the label above a bar is the number of operators
admitted, shown when at least one answerable type uses them (omitted for zero admissions and for
admitted-but-unused ones; e.g.\ 4B on Amounts \& Timing admits four operators that no answerable type ends up
using). \textbf{4B/8B/14B/32B} are Qwen3 dense models, a controlled same-family, same-generation,
same-architecture scale sweep; \textbf{35B} is the Qwen3.6-35B-A3B mixture-of-experts main model ($\approx$3B
active parameters, a later generation), shown as a cross-architecture reference point. Discussed in
Section~\ref{sec:model-ladder} and Appendix~\ref{app:model-ablation}.}
\label{fig:seed-model-ladder}
\end{figure}
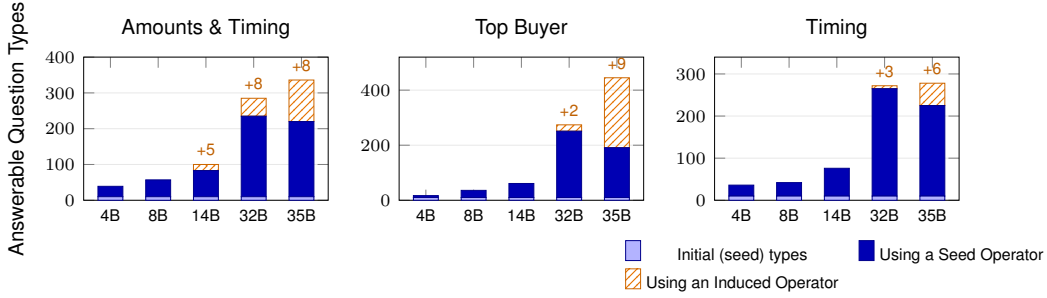

%% file: tables/base_model_ablation.tex
\begin{table}[t]
\caption{Base-model ablation on synthetic data: the full closed loop re-run with four Qwen3 dense base models on
the three starting sets and the fixed 63-merchant validation set, gate-topk $K=3$ (Appendix~\ref{app:algo}), the
same gate throughout. ``inner cov.'' is the answerable types the inner loop reaches on the seed operators; ``gap''
is the top-tallied missing capability the model discovers (each seed is given up to two induction rounds, hence
some seeds show two gaps); ``drafts'' is the operator candidates
proposed across all attempted gaps; ``admitted'' is how many pass the gate; ``final'' is the total answerable
types after all rounds for that seed. The gate is sound at every scale: \emph{no} false operator is admitted
anywhere. Self-extension is not monotone in size: 8B authors no gate-passing operator on any of its
$128$--$192$ drafts per seed despite matching or exceeding 4B's inner coverage, and 4B's own four
admitted operators unlock no further measured coverage in this run; the largest dense model, 32B, admits the most
operators overall and is the only dense model to self-extend beyond amounts/timing, on both timing and top-buyer.}
\label{tab:model-ablation}
\centering
\small\renewcommand{\arraystretch}{0.95}\setlength{\tabcolsep}{5pt}
\begin{tabular}{llccccc}
\toprule
Base model & Starting set & Inner cov. & Gap & Drafts & Admitted & Final \\
\midrule
Qwen3-4B (dense)   & Amounts/timing & 39 & group\_by, mode & 320 & 4 & 39 \\
                   & Top-buyer      & 17 & mode            & 192 & 0 & 17 \\
                   & Timing         & 36 & latest          & 128 & 0 & 36 \\
\midrule
Qwen3-8B (dense)   & Amounts/timing & 57 & mode & 128 & 0 & 57 \\
                   & Top-buyer      & 36 & mode &  64 & 0 & 36 \\
                   & Timing         & 42 & mode & 192 & 0 & 42 \\
\midrule
Qwen3-14B (dense)  & Amounts/timing & 77 & count, (unresolved) & 192 & 5 & 100 \\
                   & Top-buyer      & 61 & count               &  64 & 0 & 61 \\
                   & Timing         & 76 & mode                & 192 & 0 & 76 \\
\midrule
Qwen3-32B (dense)  & Amounts/timing & 183 & count, count\_by\_values & 256 & 8 & 285 \\
                   & Top-buyer      &  90 & count, where             & 256 & 2 & 274 \\
                   & Timing         & 137 & most\_common, max\_key    & 320 & 3 & 272 \\
\midrule
35B-A3B$^{\dagger}$ & Amounts/timing & 180 & mode, mode\_weighted &  64 & 8 & 336 \\
(main)              & Top-buyer      & 155 & count, equal         &  64 & 9 & 445 \\
                    & Timing         &  97 & mode, argmax         &  64 & 6 & 278 \\
\bottomrule
\end{tabular}

\vspace{2pt}
{\footnotesize $^{\dagger}$ Qwen3.6-35B-A3B, the mixture-of-experts model used in the main experiments
($\approx$3B active parameters, a later generation than the dense ladder); its row is the continuous
closed-loop trajectory reported in Figure~\ref{fig:trajectory}, not the dense sweep's per-seed ablation
protocol above, since (unlike the dense sweep) it is a cross-architecture reference point rather than a
controlled rung on the same axis. Both induction rounds admit operators on every seed here; ``admitted''
sums both rounds ($6{+}2$, $5{+}4$, $5{+}1$) and ``final'' is the trajectory's endpoint after both.}
\end{table}

%% file: tables/cross_family.tex
\begin{table}[t]
\caption{Cross-family summary: the full closed loop re-run with three additional families (Mistral-Small-24B,
Gemma-3-27B, Llama-3.3-70B) as the proposer, same frozen prompt and protocol as the main model, across the three
merchant seeds and two transfer domains. \emph{Final} is answerable types after both rounds; \emph{compile} is the
median fraction of proposed expressions that execute (the proposer's DSL fluency). Every family admits zero wrong
operators; discussion in \S\ref{sec:model-ladder}.}
\label{tab:cross-family}
\centering
\small\renewcommand{\arraystretch}{0.95}\setlength{\tabcolsep}{5pt}
\begin{tabular}{lccccccc}
\toprule
 & \multicolumn{3}{c}{Merchant seeds (final)} & \multicolumn{2}{c}{Transfer (final)} & False & Compile \\
\cmidrule(lr){2-4}\cmidrule(lr){5-6}
Proposer (family) & Amt & Buyer & Timing & BookSQL & EHRSQL & Admits & (median) \\
\midrule
Qwen3.6-35B-A3B (main) & 336 & 445 & 278 & 162 & 248 & 0 & n/a \\
Mistral-Small-24B      & 302 & 258 & 243 & 158 & 147 & 0 & 31\% \\
Gemma-3-27B            &  70 &  25 & 121 &  37 &  75 & 0 & 13\% \\
Llama-3.3-70B          & 155 & 143 & 181 & 109 & 106 & 0 & 34\% \\
\bottomrule
\end{tabular}

\vspace{2pt}
{\footnotesize Operators admitted per condition (amt/buyer/timing/BookSQL/EHRSQL): main $8/9/6/7/26$; Mistral
$3/7/2/1/0$; Gemma $0/0/2/0/0$; Llama $12/8/8/2/5$. Compile is the median executable fraction (executed/candidate)
over the five conditions from the committed inner-loop results; the per-pass draft-rejection analysis is in
Appendix~\ref{app:model-ablation}.}
\end{table}

%% file: tables/booksql_baseline.tex
\begin{table}[t]
\caption{Text-to-SQL baseline on a second domain, BookSQL (business-ledger accounts): 3{,}000 self-generated
compositional questions (15 templates across three categories, 200 each) over 300 accounts. The same frozen
base models emit a SQLite query (best-of-5, majority vote over the executed results, temperature 0.4, matching
the merchant baseline of Table~\ref{tab:sql} and EHRSQL's Table~\ref{tab:ehrsql}) over a \emph{flat per-account}
table that mirrors the DSL context one-to-one, so no query needs a JOIN. SQL is granted the median and
population-standard-deviation aggregates SQLite omits, defined to match the DSL. Accuracy is per query (each
(question, account) instance scored against gold), the same convention used throughout the merchant
(Table~\ref{tab:sql}) and EHRSQL (Table~\ref{tab:ehrsql}) baselines. \emph{agg} is flat aggregation,
\emph{ratio} is a derived statistic (coefficient of
variation, peak-to-average, daily volume), and \emph{order} is a consecutive-entry gap statistic, needing an
ordered window function in SQL. Numbers are the mean of 3 seeds (per-cell std $\le 0.01$); the DSL is exact by
construction. The \emph{few-shot} columns prepend three hand-authored (question, SQL) exemplars (one GROUP BY,
one ordered window, one granted aggregate), disjoint from the eval; few-shot sharply raises order for the larger
Qwen models (32B/35B to $\approx$1.00) but is model-dependent, regressing Qwen3-8B and Gemma.}
\label{tab:booksql-sql}
\centering
\small\renewcommand{\arraystretch}{0.95}\setlength{\tabcolsep}{6pt}
\begin{tabular}{lcccccc}
\toprule
 & \multicolumn{4}{c}{zero-shot} & \multicolumn{2}{c}{few-shot} \\
\cmidrule(lr){2-5}\cmidrule(lr){6-7}
model & agg & ratio & order & overall & order & overall \\
\midrule
Qwen3-4B        & 1.000 & 0.667 & 0.500 & 0.800 & 0.750 & 0.800 \\
Qwen3-8B        & 0.750 & 0.443 & 0.498 & 0.622 & 0.319 & 0.624 \\
Qwen3-14B       & 0.993 & 0.667 & 0.250 & 0.729 & 0.507 & 0.868 \\
Qwen3-32B       & 1.000 & 0.996 & 0.448 & 0.852 & 1.000 & 0.933 \\
Qwen3.6-35B-A3B & 1.000 & 0.859 & 0.514 & 0.842 & 0.990 & 0.944 \\
Gemma-3-27B     & 0.891 & 1.000 & 0.802 & 0.889 & 0.743 & 0.865 \\
\midrule
\textbf{DSL (ours)} & \textbf{1.000} & \textbf{1.000} & \textbf{1.000} & \textbf{1.000} & \textbf{1.000} & \textbf{1.000} \\
\bottomrule
\end{tabular}
\end{table}

%% file: tables/ehrsql_baseline.tex
\begin{table}[t]
\caption{Text-to-SQL baseline on a third domain, EHRSQL (MIMIC-III): the full verified pools (7{,}660
compositional and 4{,}500 retrieval instances) over 300 single-patient contexts. The same frozen base
models emit a SQLite query (best-of-5, majority vote over the executed results, temperature 0.4, matching the
merchant baseline of Table~\ref{tab:sql}) over a \emph{flat per-patient} table that mirrors the DSL context
one-to-one, so no query needs a JOIN. SQL is granted the median and population-standard-deviation aggregates
SQLite omits, defined to match the DSL. Accuracy is per query (each (question, patient) instance scored against
gold). \emph{Compositional} is the main track (self-generated statistical/ordered questions); \emph{Retrieval}
is a supplementary control of real EHRSQL attribute look-ups, where both systems are near ceiling. Numbers are
the mean of 3 seeds (per-cell std $\le 0.01$); the DSL is exact by construction. The \emph{Compositional
(few-shot)} column prepends three hand-authored (question, SQL) exemplars (one GROUP BY, one ordered window, one
granted aggregate), disjoint from the eval; it lifts most Qwen backbones (4B $+0.22$, 14B $+0.08$, 32B $+0.07$)
but is model-dependent: Gemma-3-27B \emph{drops} ($^\dagger$0.768), where the window exemplar induces an
aggregate-over-window (\texttt{AVG(LAG()\,OVER\,...)}) that SQLite rejects, a form it avoided zero-shot with
legal self-joins.}
\label{tab:ehrsql}
\centering
\small\renewcommand{\arraystretch}{0.95}\setlength{\tabcolsep}{6pt}
\begin{tabular}{lccc}
\toprule
model & Compositional (main) & Compositional (few-shot) & Retrieval (supp.) \\
\midrule
Qwen3-4B        & 0.635 & 0.852 & 0.809 \\
Qwen3-8B        & 0.763 & 0.740 & 0.775 \\
Qwen3-14B       & 0.774 & 0.850 & 0.848 \\
Qwen3-32B       & 0.821 & 0.887 & 0.808 \\
Qwen3.6-35B-A3B & 0.911 & 0.917 & 0.958 \\
Gemma-3-27B     & 0.885 & $^\dagger$0.768 & 0.859 \\
\midrule
\textbf{DSL (ours)} & \textbf{1.000} & \textbf{1.000} & \textbf{1.000} \\
\bottomrule
\end{tabular}
\end{table}

%% file: figures/convergence.tex
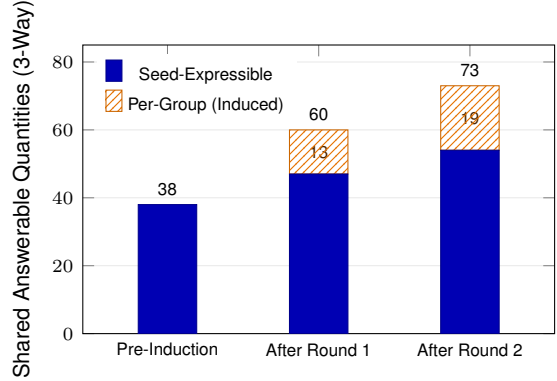
\begin{figure}[t]
\centering
\begin{tikzpicture}
\begin{axis}[
  width=0.56\linewidth, height=5.4cm,
  ybar stacked, /pgf/bar width=22pt,
  symbolic x coords={pre,r1,r2}, xtick=data,
  xticklabels={{Pre-Induction},{After Round 1},{After Round 2}},
  enlarge x limits=0.28, ymin=0, ymax=85,
  ylabel={Shared Answerable Quantities (3-Way)},
  ymajorgrids=true, grid style={gray!18, line width=0.3pt},
  tick label style={font=\sffamily\scriptsize}, label style={font=\sffamily\footnotesize},
  legend style={font=\sffamily\scriptsize, draw=none, at={(0.03,0.97)}, anchor=north west,
    legend columns=1, /tikz/every even column/.append style={column sep=6pt}}]
\addplot[fill=blue!70!black, draw=blue!55!black] coordinates {(pre,38) (r1,47) (r2,54)};
\addplot[pattern=north east lines, pattern color=orange!90!black, draw=orange!80!black] coordinates {(pre,0) (r1,13) (r2,19)};
\node[font=\sffamily\scriptsize,anchor=center,text=orange!30!black] at (axis cs:r1,53.5) {13};
\node[font=\sffamily\scriptsize,anchor=center,text=orange!30!black] at (axis cs:r2,63.5) {19};
\node[font=\sffamily\scriptsize,anchor=south] at (axis cs:pre,38) {38};
\node[font=\sffamily\scriptsize,anchor=south] at (axis cs:r1,60)  {60};
\node[font=\sffamily\scriptsize,anchor=south] at (axis cs:r2,73)  {73};
\legend{Seed-Expressible, Per-Group (Induced)}
\end{axis}
\end{tikzpicture}
\caption{Three independent starting sets converge on a shared answerable set. Bars count the quantities
answerable in \emph{all three} runs (the 3-way intersection of their answerable sets, compared by value
signature on the shared 63-merchant validation set) at each induction stage. Before any induction the three share
$38$ quantities, all expressible from their common seed operators and \emph{none} per-group. After induction the
shared core grows to $73$, of which $19$ are per-group aggregations (hatched) that lie outside every starting
set's closure, so the convergence is in the induced capability, not the seed operators. The three do not collapse
into one identical set (each also keeps exploring its own directions; final per-run sizes $326/435/268$, pairwise
overlaps $102$--$118$); rather, from disjoint seeds they independently acquire the same per-group capability and
a common set of exactly-computable quantities.}
\label{fig:convergence}
\end{figure}

%% file: tables/convergence_operators.tex
\begin{table}[t]
\caption{Convergent evolution of operators. Each row is one quantity that ends up answerable in all three runs
and evaluates to the \emph{same value} on every one of the 63 merchants (identical value signature), yet each
run computes it with a per-group operator its own outer loop authored and named independently. The starting set
fixes the path, not the destination: three disjoint seeds reinvent the same capability under different names.
Full expressions are in the committed \texttt{convergence\_results.json}.}
\label{tab:convergence-ops}
\centering
\small\renewcommand{\arraystretch}{1.05}\setlength{\tabcolsep}{5pt}
\resizebox{\columnwidth}{!}{%
\begin{tabular}{llll}
\toprule
Quantity (same value on all 63 merchants) & Amounts/timing & Top-buyer & Timing \\
\midrule
Median transactions per buyer            & \texttt{count\_per\_group} & \texttt{per\_bsigner\_count} & \texttt{mode\_counts} \\
Busiest hour's transaction count         & \texttt{count\_per\_group} & \texttt{group\_count}        & \texttt{mode\_counts} \\
Buyers with more than one transaction    & \texttt{count\_per\_group} & \texttt{group\_count}        & \texttt{mode\_counts} \\
Busiest weekday's transaction count      & \texttt{count\_per\_group} & \texttt{group\_count}        & \texttt{mode\_count\_by} \\
\bottomrule
\end{tabular}%
}
\end{table}

%% file: iclr2025_conference.bbl
\begin{thebibliography}{25}
\providecommand{\natexlab}[1]{#1}
\providecommand{\url}[1]{\texttt{#1}}
\expandafter\ifx\csname urlstyle\endcsname\relax
  \providecommand{\doi}[1]{doi: #1}\else
  \providecommand{\doi}{doi: \begingroup \urlstyle{rm}\Url}\fi

\bibitem[Cai et~al.(2024)Cai, Wang, Ma, Chen, and Zhou]{cai2024large}
Tianle Cai, Xuezhi Wang, Tengyu Ma, Xinyun Chen, and Denny Zhou.
\newblock Large language models as tool makers.
\newblock In \emph{The Twelfth International Conference on Learning
  Representations}, 2024.
\newblock URL \url{https://openreview.net/forum?id=qV83K9d5WB}.

\bibitem[Chen et~al.(1998)Chen, Cheung, and Yiu]{chen1998metamorphic}
Tsong~Y. Chen, Shing~C. Cheung, and Siu~M. Yiu.
\newblock Metamorphic testing: A new approach for generating next test cases.
\newblock Technical Report HKUST-CS98-01, Department of Computer Science, Hong
  Kong University of Science and Technology, 1998.

\bibitem[Chen et~al.(2023)Chen, Ma, Wang, and Cohen]{chen2023pot}
Wenhu Chen, Xueguang Ma, Xinyi Wang, and William~W. Cohen.
\newblock Program of thoughts prompting: Disentangling computation from
  reasoning for numerical reasoning tasks.
\newblock \emph{Transactions on Machine Learning Research (TMLR)}, 2023.

\bibitem[Ellis et~al.(2021)Ellis, Wong, Nye, Sabl{\'e}-Meyer, Morales, Hewitt,
  Cary, Solar-Lezama, and Tenenbaum]{ellis2021dreamcoder}
Kevin Ellis, Catherine Wong, Maxwell Nye, Mathias Sabl{\'e}-Meyer, Lucas
  Morales, Luke Hewitt, Luc Cary, Armando Solar-Lezama, and Joshua~B.
  Tenenbaum.
\newblock {DreamCoder}: Bootstrapping inductive program synthesis with
  wake-sleep library learning.
\newblock In \emph{Proceedings of the ACM SIGPLAN International Conference on
  Programming Language Design and Implementation (PLDI)}, 2021.

\bibitem[{Gemma Team} et~al.(2025){Gemma Team}, Kamath, Ferret, Pathak,
  Vieillard, Merhej, Perrin, Matejovicova, Ram{\'e}, Rivi{\`e}re,
  et~al.]{team2025gemma}
{Gemma Team}, Aishwarya Kamath, Johan Ferret, Shreya Pathak, Nino Vieillard,
  Ramona Merhej, Sarah Perrin, Tatiana Matejovicova, Alexandre Ram{\'e},
  Morgane Rivi{\`e}re, et~al.
\newblock Gemma 3 technical report.
\newblock \emph{arXiv preprint arXiv:2503.19786}, 2025.

\bibitem[Grand et~al.(2024)Grand, Wong, Bowers, Olausson, Liu, Tenenbaum, and
  Andreas]{grand2024lilo}
Gabriel Grand, Lionel Wong, Maddy Bowers, Theo~X. Olausson, Muxin Liu,
  Joshua~B. Tenenbaum, and Jacob Andreas.
\newblock {LILO}: Learning interpretable libraries by compressing and
  documenting code.
\newblock In \emph{International Conference on Learning Representations
  (ICLR)}, 2024.

\bibitem[Grattafiori et~al.(2024)]{llama2024herd}
Aaron Grattafiori et~al.
\newblock The {L}lama 3 herd of models.
\newblock \emph{arXiv preprint arXiv:2407.21783}, 2024.
\newblock Llama-3 architecture reference.

\bibitem[Gwet(2008)]{gwet2008}
Kilem~Li Gwet.
\newblock Computing inter-rater reliability and its variance in the presence of
  high agreement.
\newblock \emph{British Journal of Mathematical and Statistical Psychology},
  61\penalty0 (1):\penalty0 29--48, 2008.
\newblock \doi{10.1348/000711006X126600}.

\bibitem[Johnson et~al.(2016)Johnson, Pollard, Shen, Lehman, Feng, Ghassemi,
  Moody, Szolovits, Celi, and Mark]{johnson2016mimic}
Alistair E.~W. Johnson, Tom~J. Pollard, Lu~Shen, Li-wei~H. Lehman, Mengling
  Feng, Mohammad Ghassemi, Benjamin Moody, Peter Szolovits, Leo~Anthony Celi,
  and Roger~G. Mark.
\newblock {MIMIC-III}, a freely accessible critical care database.
\newblock \emph{Scientific Data}, 3\penalty0 (1):\penalty0 160035, 2016.

\bibitem[Kumar et~al.(2024)Kumar, Dibbu, Harsola, Subrahmaniam, and
  Modi]{kumar2024booksql}
Rahul Kumar, Amar~Raja Dibbu, Shrutendra Harsola, Vignesh Subrahmaniam, and
  Ashutosh Modi.
\newblock {BookSQL}: A large scale text-to-{SQL} dataset for accounting domain.
\newblock In \emph{Proceedings of the 2024 Conference of the North American
  Chapter of the Association for Computational Linguistics: Human Language
  Technologies (Volume 1: Long Papers)}, pp.\  497--516, 2024.

\bibitem[Lee et~al.(2022)Lee, Hwang, Bae, Kwon, Shin, Yang, Seo, Kim, and
  Choi]{lee2022ehrsql}
Gyubok Lee, Hyeonji Hwang, Seongsu Bae, Yeonsu Kwon, Woncheol Shin, Seongjun
  Yang, Minjoon Seo, Jong-Yeup Kim, and Edward Choi.
\newblock {EHRSQL}: A practical text-to-{SQL} benchmark for electronic health
  records.
\newblock In S.~Koyejo, S.~Mohamed, A.~Agarwal, D.~Belgrave, K.~Cho, and A.~Oh
  (eds.), \emph{Advances in Neural Information Processing Systems}, volume~35,
  pp.\  15589--15601. Curran Associates, Inc., 2022.
\newblock \doi{10.52202/068431-1134}.
\newblock URL
  \url{https://proceedings.neurips.cc/paper_files/paper/2022/file/643e347250cf9289e5a2a6c1ed5ee42e-Paper-Datasets_and_Benchmarks.pdf}.

\bibitem[McKeeman(1998)]{mckeeman1998differential}
William~M. McKeeman.
\newblock Differential testing for software.
\newblock \emph{Digital Technical Journal}, 10\penalty0 (1):\penalty0 100--107,
  1998.

\bibitem[{Meta AI}(2024)]{metallama2024llama33}
{Meta AI}.
\newblock {Llama-3.3-70B-Instruct}.
\newblock \url{https://huggingface.co/meta-llama/Llama-3.3-70B-Instruct}, 2024.
\newblock Released 2024-12-06.

\bibitem[{Mistral AI}(2025)]{mistralai2025small3}
{Mistral AI}.
\newblock Mistral {Small} 3.
\newblock \url{https://mistral.ai/news/mistral-small-3/}, 2025.
\newblock Mistral-Small-24B-Instruct-2501.

\bibitem[Novikov et~al.(2025)Novikov, V{\~u}, Eisenberger, Dupont, Huang,
  Wagner, Shirobokov, Kozlovskii, Ruiz, Mehrabian, Kumar, See, Chaudhuri,
  Holland, Davies, Nowozin, Kohli, and Balog]{novikov2025alphaevolve}
Alexander Novikov, Ng{\^a}n V{\~u}, Marvin Eisenberger, Emilien Dupont, Po-Sen
  Huang, Adam~Zsolt Wagner, Sergey Shirobokov, Borislav Kozlovskii, Francisco
  J.~R. Ruiz, Abbas Mehrabian, M.~Pawan Kumar, Abigail See, Swarat Chaudhuri,
  George Holland, Alex Davies, Sebastian Nowozin, Pushmeet Kohli, and Matej
  Balog.
\newblock {AlphaEvolve}: A coding agent for scientific and algorithmic
  discovery.
\newblock \emph{arXiv preprint arXiv:2506.13131}, 2025.

\bibitem[Qian et~al.(2023)Qian, Han, Fung, Qin, Liu, and Ji]{qian2023creator}
Cheng Qian, Chi Han, Yi~Fung, Yujia Qin, Zhiyuan Liu, and Heng Ji.
\newblock {CREATOR}: Tool creation for disentangling abstract and concrete
  reasoning of large language models.
\newblock In \emph{Findings of the Association for Computational Linguistics:
  EMNLP 2023}, pp.\  6922--6939, 2023.

\bibitem[{Qwen Team}(2026)]{qwen3.6-35b-a3b}
{Qwen Team}.
\newblock {Qwen3.6-35B-A3B}: Agentic coding power, now open to all, April 2026.
\newblock URL \url{https://qwen.ai/blog?id=qwen3.6-35b-a3b}.

\bibitem[Romera-Paredes et~al.(2024)Romera-Paredes, Barekatain, Novikov, Balog,
  Kumar, Dupont, Ruiz, Ellenberg, Wang, Fawzi, et~al.]{romera2024mathematical}
Bernardino Romera-Paredes, Mohammadamin Barekatain, Alexander Novikov, Matej
  Balog, M~Pawan Kumar, Emilien Dupont, Francisco~JR Ruiz, Jordan~S Ellenberg,
  Pengming Wang, Omar Fawzi, et~al.
\newblock Mathematical discoveries from program search with large language
  models.
\newblock \emph{Nature}, 625\penalty0 (7995):\penalty0 468--475, 2024.

\bibitem[Segura et~al.(2016)Segura, Fraser, S{\'a}nchez, and
  Ruiz-Cort{\'e}s]{segura2016survey}
Sergio Segura, Gordon Fraser, Ana~Bel{\'e}n S{\'a}nchez, and Antonio
  Ruiz-Cort{\'e}s.
\newblock A survey on metamorphic testing.
\newblock \emph{IEEE Transactions on Software Engineering}, 42\penalty0
  (9):\penalty0 805--824, 2016.

\bibitem[Wang et~al.(2024{\natexlab{a}})Wang, Xie, Jiang, Mandlekar, Xiao, Zhu,
  Fan, and Anandkumar]{wang2024voyager}
Guanzhi Wang, Yuqi Xie, Yunfan Jiang, Ajay Mandlekar, Chaowei Xiao, Yuke Zhu,
  Linxi Fan, and Anima Anandkumar.
\newblock Voyager: An open-ended embodied agent with large language models.
\newblock \emph{Transactions on Machine Learning Research}, 2024{\natexlab{a}}.
\newblock ISSN 2835-8856.
\newblock URL \url{https://openreview.net/forum?id=ehfRiF0R3a}.

\bibitem[Wang et~al.(2019)Wang, Lehman, Clune, and Stanley]{wang2019paired}
Rui Wang, Joel Lehman, Jeff Clune, and Kenneth~O Stanley.
\newblock Paired open-ended trailblazer ({POET}): Endlessly generating
  increasingly complex and diverse learning environments and their solutions.
\newblock \emph{arXiv preprint arXiv:1901.01753}, 2019.

\bibitem[Wang et~al.(2023)Wang, Wei, Schuurmans, Le, Chi, Narang, Chowdhery,
  and Zhou]{wang2023selfconsistency}
Xuezhi Wang, Jason Wei, Dale Schuurmans, Quoc Le, Ed~Chi, Sharan Narang,
  Aakanksha Chowdhery, and Denny Zhou.
\newblock Self-consistency improves chain of thought reasoning in language
  models.
\newblock In \emph{International Conference on Learning Representations
  (ICLR)}, 2023.

\bibitem[Wang et~al.(2024{\natexlab{b}})Wang, Neubig, and Fried]{wang2024trove}
Zhiruo Wang, Graham Neubig, and Daniel Fried.
\newblock {T}ro{VE}: Inducing verifiable and efficient toolboxes for solving
  programmatic tasks.
\newblock In Ruslan Salakhutdinov, Zico Kolter, Katherine Heller, Adrian
  Weller, Nuria Oliver, Jonathan Scarlett, and Felix Berkenkamp (eds.),
  \emph{Proceedings of the 41st International Conference on Machine Learning},
  volume 235 of \emph{Proceedings of Machine Learning Research}, pp.\
  51177--51191. PMLR, 21--27 Jul 2024{\natexlab{b}}.
\newblock URL \url{https://proceedings.mlr.press/v235/wang24az.html}.

\bibitem[Yang et~al.(2025)Yang, Li, Yang, Zhang, Hui, Zheng, Yu, Gao, Huang,
  Lv, et~al.]{yang2025qwen3}
An~Yang, Anfeng Li, Baosong Yang, Beichen Zhang, Binyuan Hui, Bo~Zheng, Bowen
  Yu, Chang Gao, Chengen Huang, Chenxu Lv, et~al.
\newblock Qwen3 technical report.
\newblock \emph{arXiv preprint arXiv:2505.09388}, 2025.

\bibitem[Yang et~al.(2011)Yang, Chen, Eide, and Regehr]{yang2011csmith}
Xuejun Yang, Yang Chen, Eric Eide, and John Regehr.
\newblock Finding and understanding bugs in {C} compilers.
\newblock In \emph{Proceedings of the 32nd ACM SIGPLAN Conference on
  Programming Language Design and Implementation (PLDI)}, pp.\  283--294, 2011.

\end{thebibliography}
